\documentclass[11pt]{article}

\usepackage[preprint]{acl}

\usepackage{times}
\usepackage{latexsym}

\usepackage{pgfplots}
\usepackage{subcaption}
\usepackage{enumitem}
\usepackage[compact]{titlesec}
\usepackage{xspace}
\usepackage{rotating}
\usepackage{booktabs, longtable, array, xcolor, multirow}
\usepackage{longtable}
\usepackage{array}

\definecolor{rowgray}{gray}{0.93}

\newcommand{\BENF}{\textsc{BENF}\xspace}
\newcommand{\USE}{\textsc{USE}\xspace}
\newcommand{\TRU}{\textsc{TRU}\xspace}

\newcommand{\squishlist}{
 \begin{list}{$\bullet$}
  { \setlength{\itemsep}{0pt}
     \setlength{\parsep}{2pt}
     \setlength{\topsep}{2pt}
     \setlength{\partopsep}{0pt}
     \setlength{\leftmargin}{1em}
     \setlength{\labelwidth}{1em}
     \setlength{\labelsep}{0.4em} } }

\newcommand{\squishend}{
  \end{list}  }

\usepackage[T1]{fontenc}

\usepackage[utf8]{inputenc}

\usepackage{microtype}

\usepackage{inconsolata}

\usepackage{graphicx}

\usepackage{makecell}
\usepackage{amsmath}

\usepackage[table]{xcolor}  
\usepackage{multirow}  

\usepackage{hyperref}

\title{Why Do People Turn to LLMs for Emotional Support? \\Insights from 2,400 Users Across Six Countries}

\title{The Trust Boundary: How Cultural Context Shapes LLM Adoption for Emotional Care}

\title{The Support Divide: How Cultural Factors Shape LLM Use for Emotional Wellbeing}

\title{Understanding User Perceptions of LLMs for Emotional Support: \\ A Cross-Cultural Survey Study}

\title{From Chatbots to Confidants: \\ A Cross-Cultural Study of LLM Adoption for Emotional Support}

\author{
    Natalia Amat-Lefort\thanks{\enspace Equal contribution.}\textsuperscript{1},
    Mert Yazan$^{*}$\textsuperscript{1,2},
    Amanda Cercas Curry\textsuperscript{3},
    Flor Miriam Plaza-del-Arco\textsuperscript{1} \\
    \textsuperscript{1}Leiden University \\
    \textsuperscript{2}Hogeschool van Amsterdam \\
    \textsuperscript{3}Independent Researcher \\
    \texttt{\{n.amat.lefort, f.m.plaza.del.arco\}@liacs.leidenuniv.nl} \\
    \texttt{m.yazan@hva.nl} \quad \texttt{amanda.cercas@gmail.com}
}

\begin{document}
\maketitle
\begin{abstract}

Large Language Models (LLMs) are increasingly used not only for instrumental tasks, but as always-available and non-judgmental confidants for emotional support. Yet what drives adoption and how users perceive emotional support interactions across countries remains unknown. To address this gap, we present the first large-scale cross-cultural study of LLM use for emotional support, surveying 4,641 participants across seven countries (USA, UK, Germany, France, Spain, Italy, and the Netherlands). Our results show that adoption rates vary dramatically across countries (from 20\% to 59\%). Using mixed models that separate cultural effects from demographic composition, we find that being aged 25-44, religious, married, and of higher socioeconomic status are predictors of positive perceptions (trust, usage, perceived benefits), with socioeconomic status being the strongest. English-speaking countries consistently show more positive perceptions than Continental European countries. We also collect a corpus of 731 real multilingual prompts from user interactions, showing that users mainly seek help for loneliness, stress, relationship conflicts, and mental health struggles.
Our findings reveal that LLM emotional support use is shaped by a complex sociotechnical landscape and call for a broader research agenda examining how these systems can be developed, deployed, and governed to ensure safe and informed access. We share our code and data publicly \footnote{\url{https://anonymous.4open.science/r/MentalWellbeingChatbots-2E21/README.md}}.

 
\end{abstract}

\section{Introduction}

Large Language Models (LLMs) are a jack of all trades, performing diverse tasks such as question answering and decision support \cite{chatterji2025chatgpt}. Beyond these instrumental uses, researchers and designers are increasingly interested in AI systems as potential companions or sources of interpersonal support \cite[e.g.][]{andersson2025companionship,wu-etal-2025-personas, savoldi2025generative}. In particular, conversational interfaces make LLMs accessible for emotionally oriented interactions such as seeking reassurance, discussing personal concerns, or reflecting on experiences\footnote{\url{https://hbr.org/2025/04/how-people-are-really-using-gen-ai-in-2025}}. 

\begin{figure}[t]
    \centering
    \includegraphics[width=0.95\columnwidth]{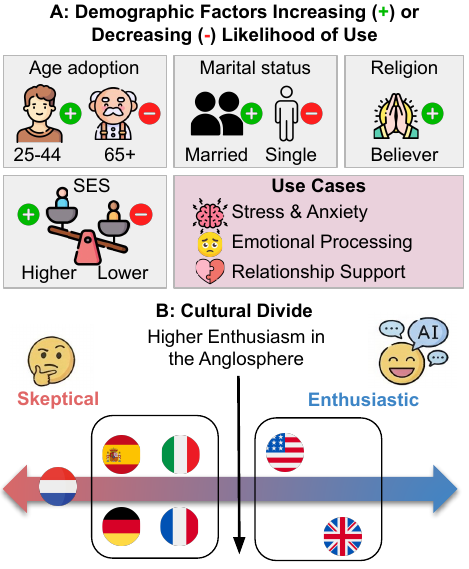}
    \caption{Global and Demographic Landscape in LLM Adoption and Perceptions for Emotional Support (N=4,641). SES: Socio-economic status.}
    \label{figure1}
\end{figure}


At the same time, many individuals face barriers to accessing traditional mental health support, including cost, stigma, and limited availability of services \cite{hou-etal-2025-language, pew2023health}. As a result, digital technologies are increasingly explored as complementary sources of emotional support \cite{chaudhry2024user, pandya2024chatgpt}. Conversational AI systems may lower barriers to seeking help by offering immediate, and always-available interaction \cite{aman2024tiktok, pandya2024chatgpt}. Yet, the growing use of LLMs for emotionally sensitive interactions raises questions about safety, trust, and appropriate system design \cite{kaffee2025intima}.


While these ethical and design implications have received significant attention, empirical evidence remains scarce on how commonly the general public turns to LLMs for emotional support and what factors shape these interactions. In particular, we lack population-level evidence on (1) who adopts LLMs for emotional support, (2) how sociodemographic and cultural backgrounds shape their perceptions of these systems, and (3) what users actually disclose to these systems. 

To address this gap, we conduct a survey examining how members of the public report using LLMs for emotional and mental health–related support. We examine whether cultural and sociodemographic factors mediate the use of LLMs for emotional support (Figure \ref{figure1}). Specifically, we address three pivotal research questions~(RQs):

\noindent
\textbf{(RQ1)}~ Who uses LLMs for emotional support, and what are the main purposes and benefits that motivate people to use them?
\\
\textbf{(RQ2)}~How do demographics and cultural background shape user perceptions of these systems?
\\
\textbf{(RQ3)}~What do users actually disclose when seeking such support?

\paragraph{Contributions} 1) We present the first cross-cultural empirical study of GenAI chatbot use for emotional support, surveying 4,641 participants across seven countries. 2) We provide an extensive sociodemographic and cultural analysis using Cumulative Link Mixed Models across three constructs (Trust and perceived privacy, Perceived benefits, and Usage intention) to disentangle demographic from cultural effects.
3) We collect and analyze 731 real user prompts across six languages, identifying key emotional support topics.

Together, our findings provide the first population-level evidence of how cultural and sociodemographic factors shape human–AI emotional support interactions, highlighting a complex sociotechnical landscape that motivates a broader research agenda to explore how these systems can be developed, deployed, and governed to promote use that is safe and informed.



\section{Related Work}

Interest in public perceptions of conversational AI for wellbeing has grown considerably, prompting surveys, usability 
studies, and social media analyses, predominantly conducted in English \cite{aman2024tiktok, 
chaudhry2024user, cameron2018assessing}. Findings suggest that users value the availability, non-judgmental nature, and accessibility of AI conversational agents, while pointing to shallow and context-insensitive responses as key limitations \cite{chaudhry2024user, aman2024tiktok}. 
Younger adults and those with higher education and AI familiarity tend to show relatively more openness toward AI tools 
\cite{pew2023health}. These attitudinal patterns are consistent with the CASA paradigm, which predicts that people apply the same social norms to AI as 
to humans \cite{nass2000machines} and disclose more under 
conditions of reduced social presence \cite{papneja2025self}. 
The psychological barriers underlying AI resistance, including perceived opacity, emotionlessness, and rigidity, further shape whether individuals are willing to engage with these systems \cite{defreitas2024psychological}. 
Beyond wellbeing, broader surveys of general LLM use show that SES and demographic factors stratify adoption rates and interaction styles, with higher-SES users interacting more abstractly \cite{bassignana-etal-2025-ai} and younger, more educated users showing higher adoption and literacy \cite{savoldi2025generative}.

Yet none of these works study the adoption and user perceptions of LLMs specifically for emotional wellbeing support, and all remain confined to a single country and predominantly Anglophone populations, leaving open whether adoption and perception patterns vary across cultural and sociodemographic contexts. 
We address this gap by surveying 4,641 participants across seven countries in six languages, collecting prompts from real prior interactions, and providing the first cross-cultural analysis of LLM adoption for emotional support that explicitly separates demographic from cultural effects. 




\section{Survey Methodology}

\subsection{Study Design}



We conduct a cross-cultural survey to examine demographic and cultural variation in the adoption and perception of LLMs for emotional support. We select seven countries representing the highest global shares of ChatGPT\footnote{Currently the most used conversational LLM.} visitors \cite{firstpagesage2026chatgpt}: the United States (17.1\%), France (4.3\%), Spain (3.7\%), the United Kingdom (2.7\%), Italy (2.5\%), Germany (2.4\%), and the Netherlands (1.1\%), ensuring our sample reflects the populations most actively engaging with these systems. Between November 11 and 22, 2025, we collected the data via a Qualtrics panel,\footnote{Qualtrics provides reliable access to diverse, cross-cultural respondents and robust built-in data quality checks.} using quota sampling to ensure even distribution across countries (N=200 per country) and gender balance within each national cohort.


Before starting the survey, we presented participants with an onboarding screen to obtain informed consent. To establish a shared baseline understanding, we provided examples of ``AI agents'' (e.g., ChatGPT, Gemini, Claude) and defined ``emotional support and mental wellbeing'' as using these tools for managing stress, processing emotions, or personal reflection. We informed participants of the survey's anonymous, voluntary nature, the estimated duration, and their right to withdraw at any time. The survey then proceeded in four steps. First, participants completed a demographic questionnaire. Second, a filter question classified them as \textit{users} (having used AI chatbots for emotional support) or \textit{non-users}, enabling demographic comparison between adopters and non-adopters.
Third, identified users reported their usage patterns (\S\ref{sec:usage_quest}) and evaluated their perceptions 
across several constructs. 
Finally, the survey concluded with an optional open-ended text where users could share their most recent support-seeking prompts. To maintain data privacy, we instructed participants to exclude any names, personal details, or identifiable information before submitting their text.

\paragraph{Pilot test} 
A cross-disciplinary panel of 20 experts from our target countries—including specialists in survey methodology, NLP, and HCI—reviewed and refined the survey instrument. The original English survey was automatically translated into Spanish, French, Italian, German, and Dutch via Qualtrics and subsequently reviewed by native-speaking experts from the panel to ensure linguistic and cultural accuracy. Pilot feedback informed refinements to question wording; for instance, absolute statements were softened to better capture user perception (e.g., changing ``I can express my feelings'' to ``I feel that I can express my feelings''). The average completion time of 8-9 minutes was used to determine participant compensation on the platform.

Following the pilot, we deployed the survey and collected 5,319 responses. To ensure data integrity, we applied Qualtrics quality filters to exclude suspected bot activity, duplicate submissions, incomplete or straight-line responses. In addition, we excluded participants with completion times faster than 30\% of the median, and those who failed either of two embedded attention checks. The final validated sample consisted of 4,641 participants.

\subsection{Questionnaire Design}

The survey comprises four sections: demographics, AI usage, evaluation of user perception constructs, and prompt collection.
\paragraph{Demographics} \label{sec:demo_quest}

Participants reported age, gender, nationality, education, religion, and marital status. We assessed subjective socioeconomic status using the MacArthur Scale of Subjective Social Status 
\cite{adler2000relationship}, which captures perceived social standing rather than objective income, making it particularly suited for cross-cultural comparison. We handled all data in compliance with GDPR and ensured that participants were fully anonymized. Further details are provided in \hyperref[sec:ethics]{Ethical Considerations}.

\paragraph{Usage Patterns} \label{sec:usage_quest}
 We asked users about their AI usage patterns, including the AI agents used, 
 frequency of use, primary devices, and their specific emotional support use cases.

\subsection{Constructs to Capture User Perceptions} \label{sec:dimensions}

We assess three key areas to capture user perceptions of LLMs for emotional support. To create robust, composite response variables for our Cumulative Link Mixed Model (CLMM) analysis, we merged conceptually related questionnaire sub-scales into three distinct constructs. Each item was measured on a 5-point Likert scale (from \textit{Strongly disagree} to \textit{Strongly agree}). The full questionnaire, including items drawn from existing validated scales and those developed by the authors, is provided in Table~\ref{tab:survey_items} in Appendix \ref{app:quest}. The three resulting constructs are:

\squishlist


    
    

    \item \textbf{Trust and Privacy (\TRU):} This construct merges the ``Trust'' and ``Privacy'' sub-scales to capture the user's overall confidence in the system. It assesses the agent's perceived credibility, honesty, and transparency, alongside the user's perceived safety regarding the handling of sensitive data, interaction confidentiality, and the clear communication of system limitations \cite{marimon2024, schmidmaier2024}.

    \item \textbf{Perceived Benefits (\BENF):} This construct evaluates the perceived advantages of using LLMs for emotional support. It captures practical, low-barrier benefits such as 24/7 accessibility and cost-effectiveness, alongside psychological benefits, including the ability to engage in non-judgmental exploration, gain new perspectives, and receive actionable advice.
    

    \item \textbf{Usage Intention and Behavior (\USE):} To measure overall system engagement, we merged the ``Attitude Toward Using'' and ``Actual System Use'' sub-scales, which are grounded in the Technology Acceptance Model (TAM) \cite{davis1989}. This combined construct captures both the user's favorable evaluation of the LLM as a supportive tool and their self-reported frequency of behavioral engagement.
    






\squishend

\section{Results}

We present findings along our three RQs.



\subsection{User and Non-User Profiles} \label{sec:UserVsNonUser}
\label{Demographics}

To address \textit{RQ1} 
we first compare demographic profiles of users against non-users. We then examine the users interaction patterns and underlying motivations.

\begin{figure}[t]
    \centering
    \includegraphics[width=\columnwidth]{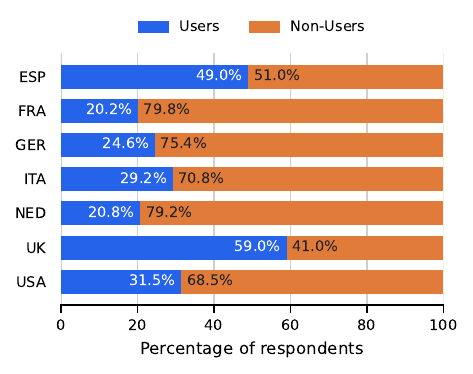}
    \caption{Cross-country comparison of AI chatbot use for emotional support and mental wellbeing.} 
    \label{fig:UserVsNonUser}
\end{figure}

\begin{figure}[ht]
    \centering
    \includegraphics[width=0.8\columnwidth, trim=0 0 0 120, clip]{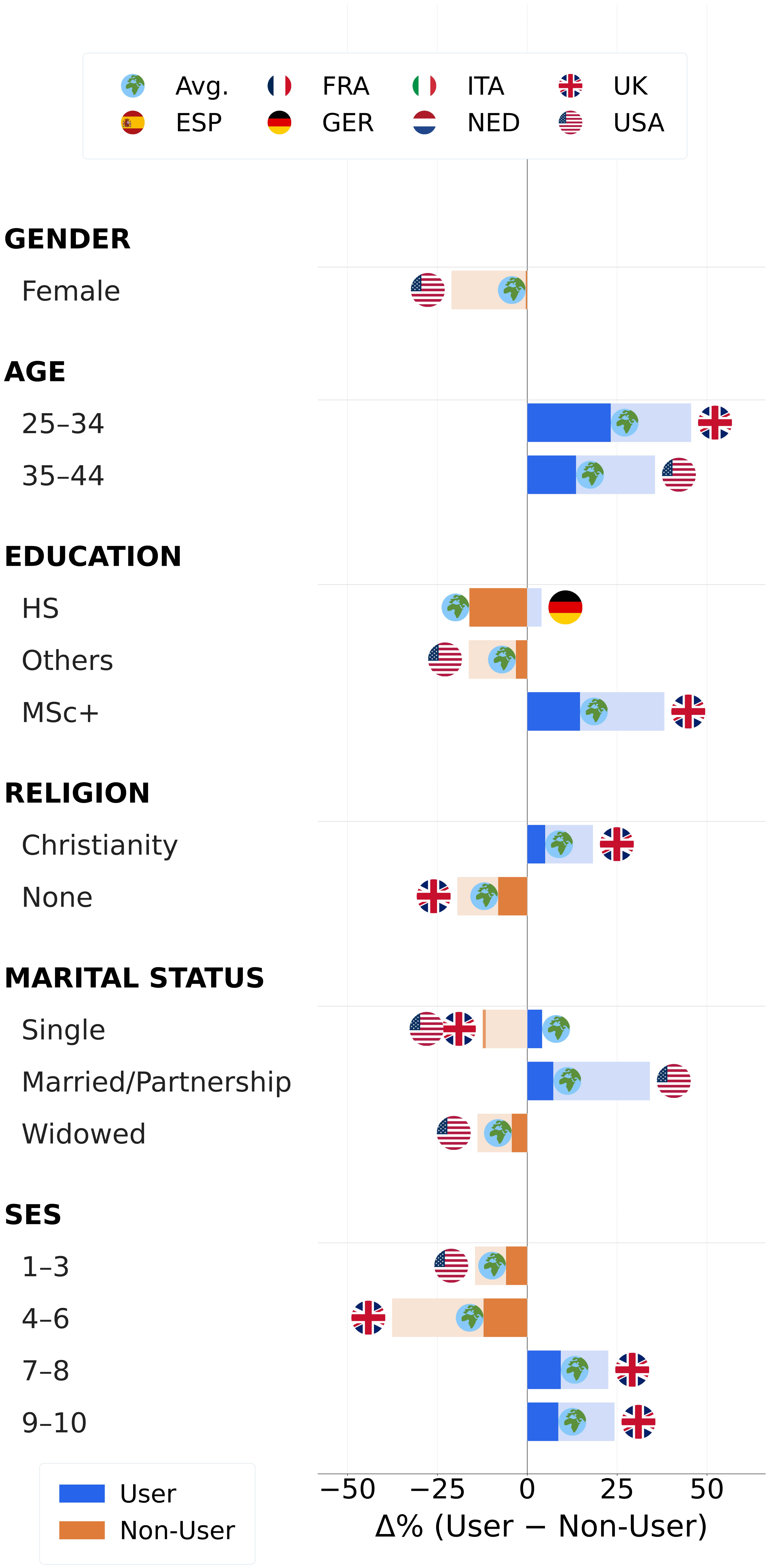}
    \caption{Percentage-point differences in demographic composition between users and non-users. Tinted bars indicate countries deviating $>$1.5 SD from the global average (marked with the globe emoji).}
    \label{fig:UserGroupDiff}
\end{figure}

\begin{figure*}[t]
    \centering
    \includegraphics[width=\textwidth]{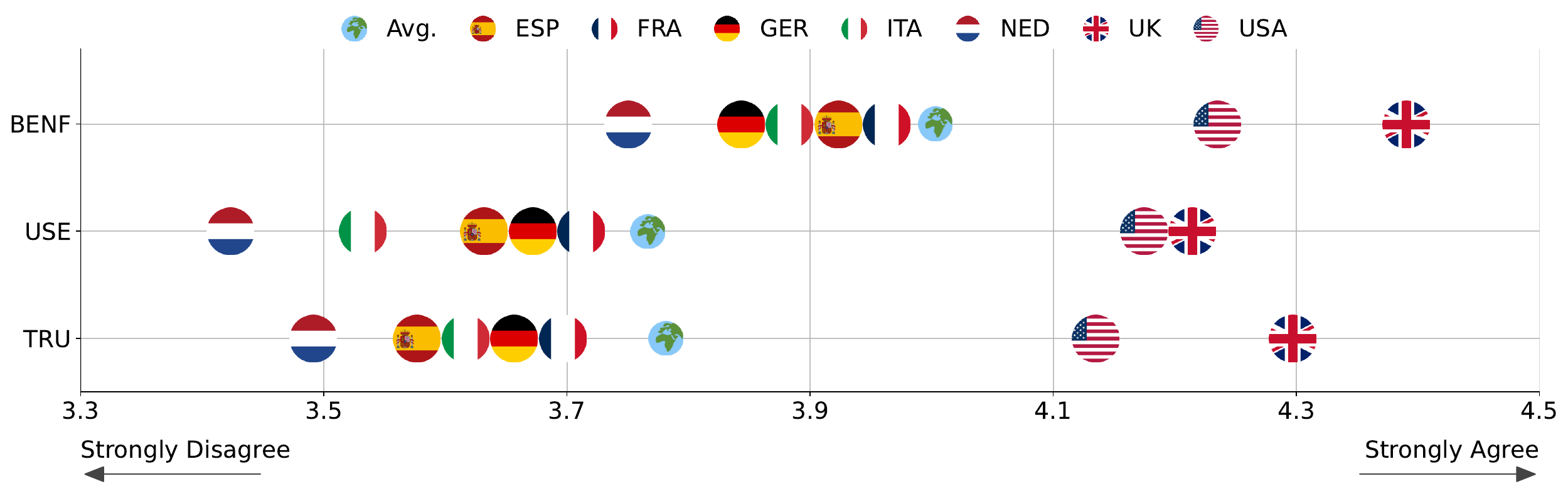}
    \caption{Mean scores per construct: Perceived Benefits (\BENF), Usage intention (\USE), Trust \& Privacy (\TRU) across seven countries in a 5-point Likert scale (from Strongly disagree to Strongly agree).}
    \label{fig:ConstructByCountry}
\end{figure*}

\paragraph{Adoption Rates by Country} Figure ~\ref{fig:UserVsNonUser} shows the distribution of users and non-users 
per country. We observe noticeable differences between the countries. The only country in which users constitute the majority is the United Kingdom (59.0\%). Spain comes close with 49.0\% users. France, Germany, and the Netherlands report the lowest percentage of users (approximately 20\% of participants). Across the seven countries surveyed, an average of 28.9\% of participants are users.

\paragraph{Aggregate Demographic Differences} To investigate demographic differences between users and non-users, we conducted chi-square tests. Except for gender ($\chi^2(4) = 3.4$, $p_{\text{adj}} = .495$), all demographic variables show significant differences: age ($\chi^2(6) = 915.3$, $p_{\text{adj}} < .001$), education ($\chi^2(7) = 319.0$, $p_{\text{adj}} < .001$), SES ($\chi^2(9) = 212.9$, $p_{\text{adj}} < .001$), marital status ($\chi^2(5) = 103.3$, $p_{\text{adj}} < .001$), and religion ($\chi^2(8) = 72.5$, $p_{\text{adj}} < .001$). Effect sizes range from small to moderate, with age showing the largest difference between the groups.


\paragraph{Demographic Profiles and Country-Level Variation}
Figure~\ref{fig:UserGroupDiff} shows the difference in demographic distribution across user groups. Users are predominantly aged 25-44, highly educated (MSc, PhD), report higher socioeconomic status, and are more likely to be married or partnered. Non-users are more likely to be individuals whose highest attained education is high school, single or widowed, and non-religious. Notable country-level differences emerge particularly in the UK and the USA. The UK shows the starkest age, education, and SES gaps, with younger and higher-SES individuals more prevalent among users, who are also more Christian and less non-religious than non-users. In the USA, non-users skew female, are overrepresented in the 35-44 age range, are more likely to be married, and are less likely to have lower SES. Germany is an exception where high-school graduates are more prevalent among users. For full demographic breakdowns, see Table~\ref{tab:UserNonUser_Demographics} and Table~\ref{tab:Demographics_ByCountry} in Appendix~\ref{sec:app_demographics}.

\paragraph{Interaction Patterns and Purposes} Moving beyond demographics, we examined how users interact with these systems (see Figure \ref{fig:usagePatterns} in Appendix \ref{app:appa}). Usage is roughly split between occasional users (34.0\%) and frequent, weekly users (31.5\%). ChatGPT is the dominant AI agent (79.7\%), and smartphones are the preferred interface by a wide margin (88.2\%). Users primarily turn to these systems for the following purposes: managing stress and anxiety (49.4\%), finding encouragement or motivation (42.5\%), processing and reflecting on emotions (41.3\%). A notable share also report experimenting out of curiosity (44.9\%), suggesting that for many, intentional support-seeking and exploratory use co-occur. Other common uses of personal support include relationship support (30.2\%) and companionship to combat loneliness (17.6\%).

\paragraph{Perceived Benefits} Figure \ref{fig:topBenefits} shows that users primarily value 
accessibility and practical utility. The highest-rated benefits are 24/7 availability (mean score = 4.26) and cost-effectiveness (4.12), indicating that users highly value immediate, low-barrier access to support. Additionally, psychological safety emerged as a critical advantage, with users highly rating the ability to express feelings without fear of human judgment (4.10).

In summary, LLM adoption for emotional support skews toward younger, higher-educated, and higher-SES individuals. Users are primarily motivated by stress management, emotional processing, and encouragement, drawn by the 24/7 availability, cost-effectiveness, and non-judgmental nature of these systems.

\begin{figure*}[t]
  \centering
  \includegraphics[width=0.8\textwidth]{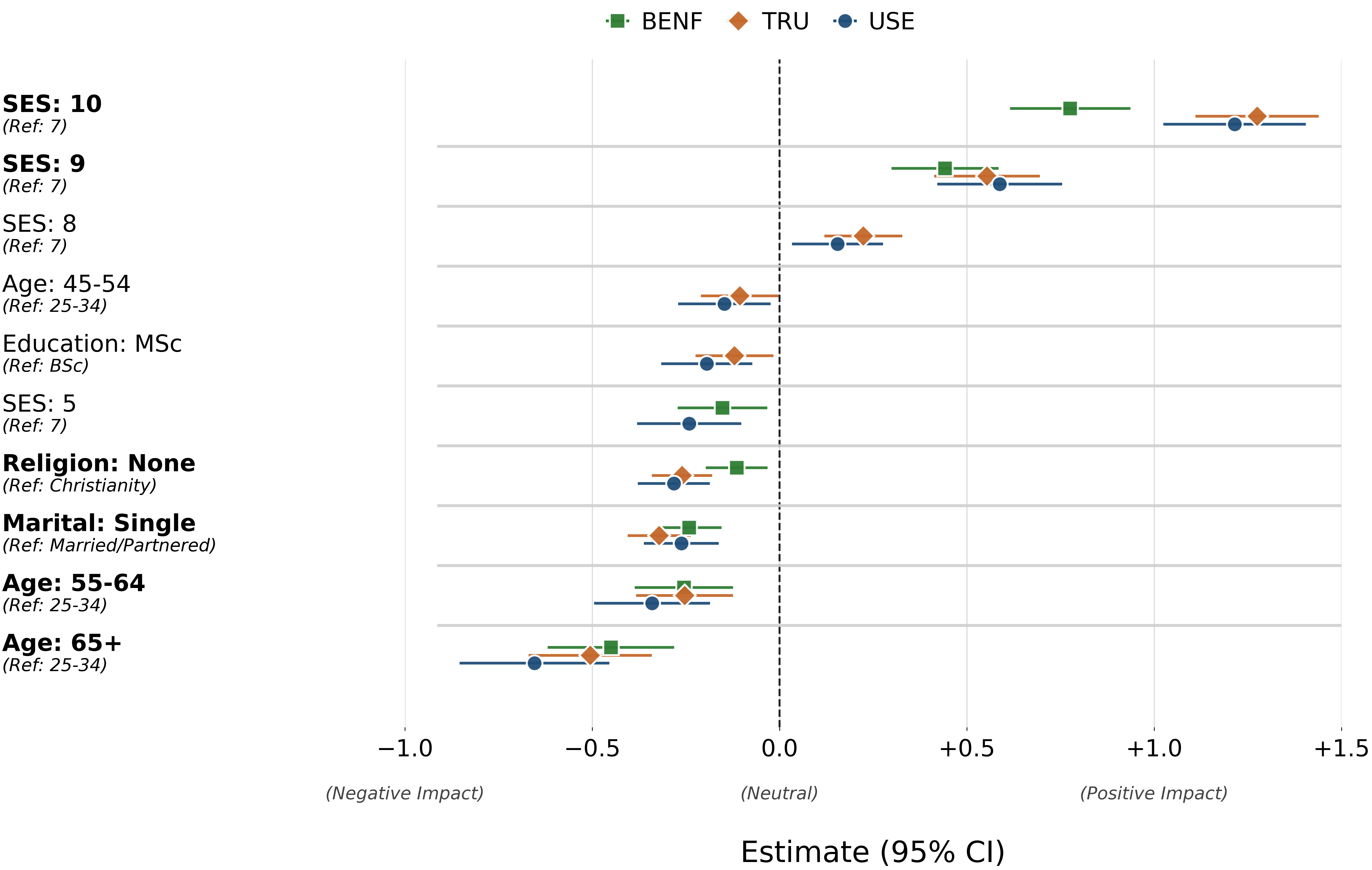}\hfill
  \caption{Mixed-model estimates of demographic factors (fixed effects) that are significant in at least 2 of the 3 constructs (BENF, USE, TRU). The ones that are significant in all three are in bold. The positive and negative impacts depend on the reference category.}
  \label{fig:CLMM_Forest}
\end{figure*}

\subsection{Cross-Country Variations in Perception}
\label{culture_constructs}

To address \textit{RQ2}, 
we first investigate how these perceptions vary across different cultural backgrounds at the country level.

Figure~\ref{fig:ConstructByCountry} presents the mean scores per construct across seven countries on a 5-point Likert scale. There is a sharp divide between the US and the UK, and the other countries. The \textbf{UK} has the highest mean scores across all three constructs, followed by the \textbf{USA}. The Anglosphere countries are above the combined mean, while mainland European countries consistently score lower. Within Europe, the \textbf{Netherlands} stands out by scoring lowest across the three constructs. To confirm statistical significance, we apply the Kruskal-Wallis test followed by Dunn's test to identify pairwise differences. The UK and the USA stand apart from all other countries across all constructs, and differ only on Perceived Benefits (\BENF). Mainland European countries are largely homogeneous: Germany, France, and Spain show no significant differences within themselves. Italy also does not differ from Spain and Germany. The Netherlands stands out among European countries: it diverges from France on all constructs, and from Germany and Spain on two constructs each. For more details, see Appendix~\ref{DescStatsConstructs}.

\begin{figure*}[t]
  \centering
  \includegraphics[width=0.8\textwidth]{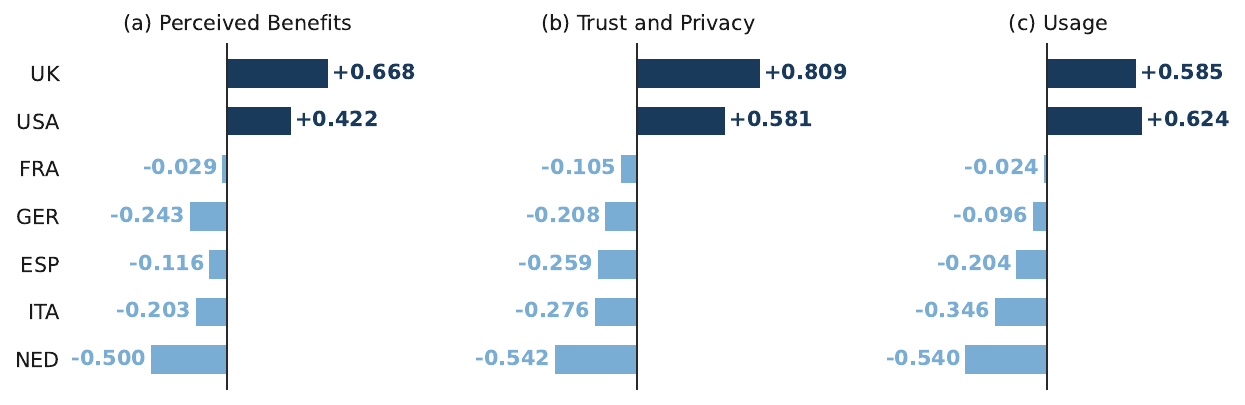}
  \caption{Country intercepts (random effects) between the three models (BENF, USE, TRU). Being from the UK and the USA is associated with positive perceptions, while mainland European countries have more negative baselines.}
  \label{fig:Intercepts}
\end{figure*}

\subsection{Mixed Model Analysis of Culture and Demographics}
\label{MixedModel}

Our findings in \S\ref{culture_constructs} show that perceptions vary significantly across countries. However, according to \S\ref{Demographics}, the demographic composition across countries is not necessarily similar. For example, UK and US participants are predominantly younger, highly educated, report higher SES, are more religious, and have higher marriage rates (see Figure~\ref{fig:UserGroupDiff}). Therefore, in this section, we answer \textit{RQ2} by examining whether observed cross-country differences reflect genuine cultural effects or are an artifact of demographic composition.

\paragraph{Methodology}

We use a Cumulative Linear Mixed Model (CLMM) to predict single constructs with demographic variables as fixed effects and country as a random effect. We excluded demographic categories with too few samples ($<50$) and collapsed SES scores 1–3 into a single lowest category due to small cell sizes. The most populous category within each variable is selected as the reference group. We report results for three models that estimate demographic and cultural effects on Perceived Benefits (BENF), Trust \& Privacy (TRU), and Usage Intention (USE), respectively.

\paragraph{Results}


The full result tables for all three models are presented in Appendix~\ref{App_MixedModelConstructs}. Figure~\ref{fig:CLMM_Forest} provides a visual outline of the fixed effects that are significant in at least two of the models. 
We find that SES does not linearly influence perceptions, but rather, being in the highest group (>7) has a noticeable impact. Similarly, being older than 45 lowers perceptions exponentially, and being 65+ is the most negative effect across all three models. However, being younger (18-24) does not consistently have a positive influence. 

Religion shows steady findings across models, with religious individuals having more positive perceptions. The differences between Christianity, Islam, and other religions are insignificant compared to the divide between believers and non-believers, suggesting that specific religious affiliation matters less than whether someone holds a religious belief. Similarly, marital status is consistent with married or partnered individuals showing more positive perceptions than singles across all models. However, there isn't a significant difference between married/partnered and divorced individuals. We observe that gender does play a significant role in only one construct (trust and privacy), with a small effect size. The education group also provides inconsistent findings. MSc graduates (compared to BSc) have more negative perceptions in two of the three models. However, the effect sizes are small, and we don't see any other differences between education categories.

The Intraclass Correlation Coefficient (ICC), which quantifies how much of the variance is attributable to random effects, is 4.2\% for BENF, 6.1\% for TRU, and 5.0\% for USE. This suggests that demographics are the primary drivers of variation in perceptions. Likelihood ratio tests comparing the full mixed models against fixed-effects-only specifications confirm that the country-level random effect is statistically significant across all models (p < .001), indicating that meaningful between-country variation persists even after accounting for demographic composition. Figure~\ref{fig:Intercepts} shows that across all three models, the UK and the US have large, positive intercepts. Meanwhile, mainland European countries have negative ones. The Netherlands consistently shows the largest negative effect, in line with previous findings (see Figure~\ref{fig:ConstructByCountry}). Besides that, compared to other countries, France is more neutral, with intercepts closer to zero.

In summary, \textbf{both demographics and cultural background significantly shape user perceptions, though demographics account for the majority of the variance}. Demographically, high socioeconomic status is the strongest driver of positive perceptions, followed by being 25-45, religious, and married or partnered. Culturally, even after controlling for demographic differences, a significant geographic divide remains: users in the Anglosphere (the UK and the USA) hold notably positive views toward using LLMs for emotional support, whereas mainland European countries display more negative perceptions, with the Netherlands being the most skeptical.

\subsection{Prompt Analysis}\label{sec:prompt_analysis}


\paragraph{Methodology}
To address \textit{RQ3}, we analyze the prompts shared by participants during the survey. Of the 1,343 users, 54.4\% opted to share their most recent emotional support message (n=731). Sharing rates varied considerably across countries: the UK was most willing (67.4\%), followed by Italy (62.1\%), France (58.9\%), the USA (58.2\%), and the Netherlands (50.8\%), while Spain (46.5\%) and Germany (38.5\%) shared the least. To identify categories, we cluster the prompts with HDBSCAN \cite{hdbscan} and UMAP \cite{umap}. We embedded the prompts using SentenceTransformer \cite{reimers-gurevych-2019-sentence} and the multilingual M3-Embedding \cite{chen-etal-2024-m3}. Clusters are assigned a category through a combination of GPT-5.2 \cite{openai2025gpt52} and manual evaluation.  After identifying mental wellbeing prompts, we applied the same clustering logic to analyze the underlying topics within this subset.

\paragraph{Results} 
Of the 731 shared prompts, 267 (36.5\%) are about mental wellbeing and interpersonal connection. The remainder covered physical health and bodily autonomy (n=93), productivity and knowledge (n=158), and others, such as filler text or refusals (n=213). 
Within the wellbeing and interpersonal connection clusters, we identified five key topics: general stress and anxiety, relationship conflict, family and caregiving, loneliness, and trauma and mental health. Topic distributions are consistent across countries (see Figure~\ref{fig:PromptTopics}). Examples of shared prompts representing these key topics are shown in Table~\ref{tab:prompts_by_country} in Appendix~\ref{PromptSamples}.

\section{Discussion}




Our results show heterogeneity along geographic and sociodemographic lines: Demographics play a significant role in shaping perceptions, with SES and age group emerging as the most influential determinants. While Anglosphere users present high trust, mainland European users remain skeptical regarding how LLMs handle sensitive information. Perceptions are particularly negative in the Netherlands, consistent with previous findings showing Dutch citizens rank last globally in attitudes toward AI \cite{dutch_ap_2025_report}. This is due 
to a complex interaction between mental health stigma, access to care, digital literacy, and privacy concerns.

Regarding \textit{mental health stigma}, ``psychological safety'' is a key driver of adoption: users readily disclose sensitive information because they do not fear human judgment, 
consistent with the model of computer-mediated communication 
\cite{joinson2001self}. Users are not just using LLMs as a cheaper substitute for human therapists; they are seeking them out \emph{because} they are not human. However, this perceived safety may be partly illusory: LLMs systematically over-affirm users rather than offering critical engagement \cite{cheng2025elephant}. 

\textit{Mental health stigma} further explains country-wise differences:
Participants in the UK are the most common adopters of LLMs for emotional wellbeing and report more positive perceptions. Notably, adoption rates are much higher than in the US despite similar ratings across constructs, possibly reflecting differing attitudes towards mental health support, as it is more stigmatised in the U.K. \cite{digiuni2013perceived, todd1974us}. Another possible explanation is \textit{access}. The UK offers accessible, free-at-the-point-of-use care through the National Health Service (NHS), but with significant wait times. Therefore, Britons may perceive LLMs as beneficial for emotional support. 
Similarly, mental health stigma in Spain has significantly decreased over the last few years \cite{varaona2024snapshot}, but mental health support access remains low in many European countries \cite{fiorillo2025roadmap, oblak2025public}. 

Our results broadly align with general adoption patterns for LLMs: their use is more common among younger and richer people \cite{bassignana-etal-2025-ai}. These groups are generally considered to have higher digital literacy. However, \textit{digital literacy} involves not only the ability to use technologies, but to engage with them critically.\footnote{\url{https://www.cedefop.europa.eu/en/tools/vet-glossary/glossary/digitale-alphabetisierung}} Given the uncertain benefits and dangers of relying on such systems in emotionally sensitive contexts, as well as the potential \textit{privacy implications} of sharing personal information with them, it is not clear that this practice necessarily reflects digital literacy. Dutch Data Protection Authority (AP) warned that commercial AI therapy apps frequently harvest sensitive data \cite{dutch_ap_2025_report}. Furthermore, OpenAI's CEO acknowledged the absence of legal ``doctor-patient confidentiality'' when using chatbots as therapists \cite{techcrunch_altman_2025}. While users feel safe, their data remains legally unprotected.

Beyond privacy concerns, emotional interaction with AI systems raises important ethical concerns \cite{cercas-curry-cercas-curry-2023-computer,gabriel2024ethics}. Even among countries with lower adoption rates, 1 in 5 people report getting emotional support from LLMs. However, we currently lack benchmarks and guidelines regarding desirable LLM behaviour in emotional support roles. Existing work raises serious concerns: LLMs consistently prioritize companionship-reinforcing over boundary-maintaining behaviours \cite{kaffee2025intima}, foster dependency with documented risks of self-harm in vulnerable users \cite{chu2025illusions}, and, despite displaying empathy, fail to interpret emotions genuinely and can make value judgements about certain identities \cite{cuadra2024illusion}. These limitations are compounded by a broader structural gap: emotion modeling in NLP has largely assumed emotions are universally experienced, failing to account for demographic and cultural variation \cite{plaza-del-arco-etal-2024-emotion}, and NLP systems further misascribe emotions based on stereotypes \cite{plaza-del-arco-etal-2024-angry, plaza-del-arco-etal-2024-divine}. This structural gap also extends to language. As \cite{guo2025large} demonstrate, even multilingual LLMs show English-centric biases that result in unnatural vocabulary and syntax when generating text in non-English languages. Consequently, the Anglosphere enthusiasm we observe may simply reflect a superior user experience for English-speaking users.

Together, our findings point to a widening gap between the perceived emotional competence of LLMs and their demonstrated limitations in sensitive interactions. 
As adoption grows, benchmarks, governance frameworks, and AI literacy programmes that help users critically assess these systems become increasingly urgent.


\section{Conclusion}


We survey the use of LLMs for emotional and mental wellbeing support across seven countries, examining how cultural and sociodemographic factors shape adoption and user perceptions. We find statistically significant differences in both adoption rates and perceptions across all measured constructs. In particular, higher socioeconomic status, younger age, being partnered, and religious belief are consistently associated with more positive perceptions of trust, perceived benefits, and usage intention. We collect 731 prompts from real prior interactions, finding that use centers on stress, anxiety, loneliness, and relationship difficulties. Our cross-cultural analysis reveals a robust Anglosphere–Continental Europe divide that persists even after controlling for demographics. 
Future work should explore adoption of these tools beyond Western contexts, develop multilingual benchmarks for emotionally sensitive interactions, and establish governance frameworks and AI literacy programmes to ensure safe and informed use.

\section*{Limitations}

Several limitations should be acknowledged. First, regarding sample representativeness: demographic composition varied across countries in age, education, and socioeconomic status, and within countries, the demographic profiles of AI users and non-users diverged notably, particularly in the UK and the USA. Additionally, panel respondents may differ from the general population in digital literacy. Second, the user/non-user classification relied on a single self-report filter question, which may be subject to social desirability bias or varying interpretations of what constitutes "emotional support" across languages and cultures. Third, the seven-country sample is limited to Western and Southern European contexts plus the USA, which may limit generalizability to regions where norms around self-disclosure, mental health, and technology use differ. Fourth, Likert-scale self-reports are susceptible to acquiescence bias and cultural response styles, which may partially confound cross-cultural comparisons. 

\section*{Ethical Considerations} 
\label{sec:ethics}

The present study was approved by the Institutional Ethics Board\footnote{Institution name withheld for blind review}. 
Research on the use of LLMs for emotional support and mental wellbeing raises ethical considerations related to privacy, mental health sensitivity, and responsible interpretation of user experiences. Because individuals may disclose highly personal information when interacting with AI systems in emotionally vulnerable contexts, careful attention to data collection, participant protection, and reporting practices is required.

Users may discuss mental health struggles, interpersonal conflicts, or other highly personal topics when interacting with LLMs. Collecting descriptions or examples of such interactions therefore poses risks of exposing sensitive information. To mitigate this, we did not collect directly identifying information and asked participants not to include identifying details in open-ended responses or example prompts. Data were analyzed in aggregated form, and any illustrative examples were edited or paraphrased to prevent re-identification. More broadly, our findings highlight a tension between the perceived privacy of LLM interactions and the potential sensitivity of the information users disclose to AI systems.

Participants were informed about the purpose of the study, the types of information collected, including sociodemographic variables and descriptions of LLM use, and how their responses would be used for research and publication. Participation was voluntary, and participants could withdraw before submitting the survey.

Although the study does not target clinical populations, individuals seeking emotional support from LLMs may be experiencing distress or limited access to other forms of support. Survey questions were therefore designed to focus on patterns of use and perceptions (e.g., acceptance of and anthropomorphic attitudes toward LLMs) rather than encouraging detailed disclosure of traumatic experiences. When reporting results, we avoid framing LLMs as substitutes for professional mental health care and instead emphasize that the study captures users’ perceptions and behaviors.

To minimize ambiguity and avoid potentially leading or stigmatizing questions about mental health, the survey relied on established measurement scales for constructs such as technology acceptance and anthropomorphic perceptions of AI. Using validated instruments helps ensure that questions are interpretable, comparable to prior research, and less likely to introduce unintended framing effects when studying sensitive behaviors related to emotional support. 

 Participants received compensation aligned with the norms of the recruitment platform and the estimated survey duration. 



\section*{Acknowledgements}
We would like to thank Joost Visser, Joost Broekens, Suzan Verberne, Elisa Bassignana, Frederic Marimon, Marta Mas-Machuca, Kudzai Sauka, Vera Neplenbroek, Roxana Petcu, and
Mohanna Hoveyda. Their feedback and insights played a key role in refining our questionnaire and shaping the methodology of this work. We also gratefully acknowledge the financial support provided by the Human AI cluster and the Software, AI \& Business cluster at the Leiden Institute of Advanced Computer Science (LIACS).

\bibliography{custom}

\appendix

\clearpage
\section{User Interaction Patterns and Perceived Benefits}\label{app:appa}

\begin{figure*}[t]
    \centering
    \includegraphics[width=0.9\textwidth]{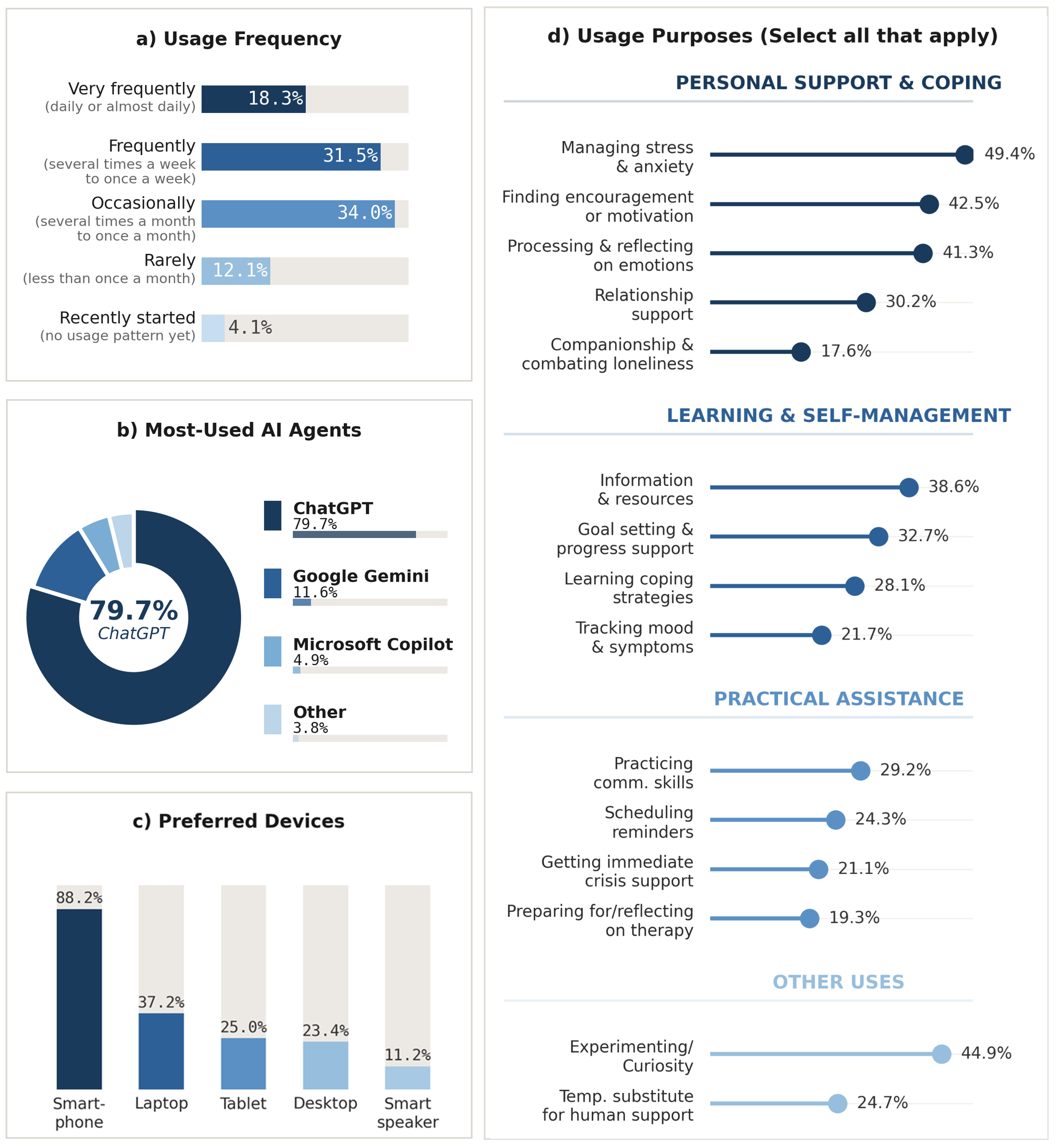}
    \caption{Overview of user interaction patterns with AI agents for mental wellbeing and emotional support: distribution by (a) Frequency of use, (b) Most-used agents, (c) Preferred devices, and (d) Grouped functional purposes.}
    \label{fig:usagePatterns}
\end{figure*}

Figure \ref{fig:usagePatterns} presents an overview of user interaction patterns with AI agents for mental wellbeing and emotional support, including the main usage purposes. 

\begin{figure*}[t]
    \centering
    \includegraphics[width=0.8\textwidth]{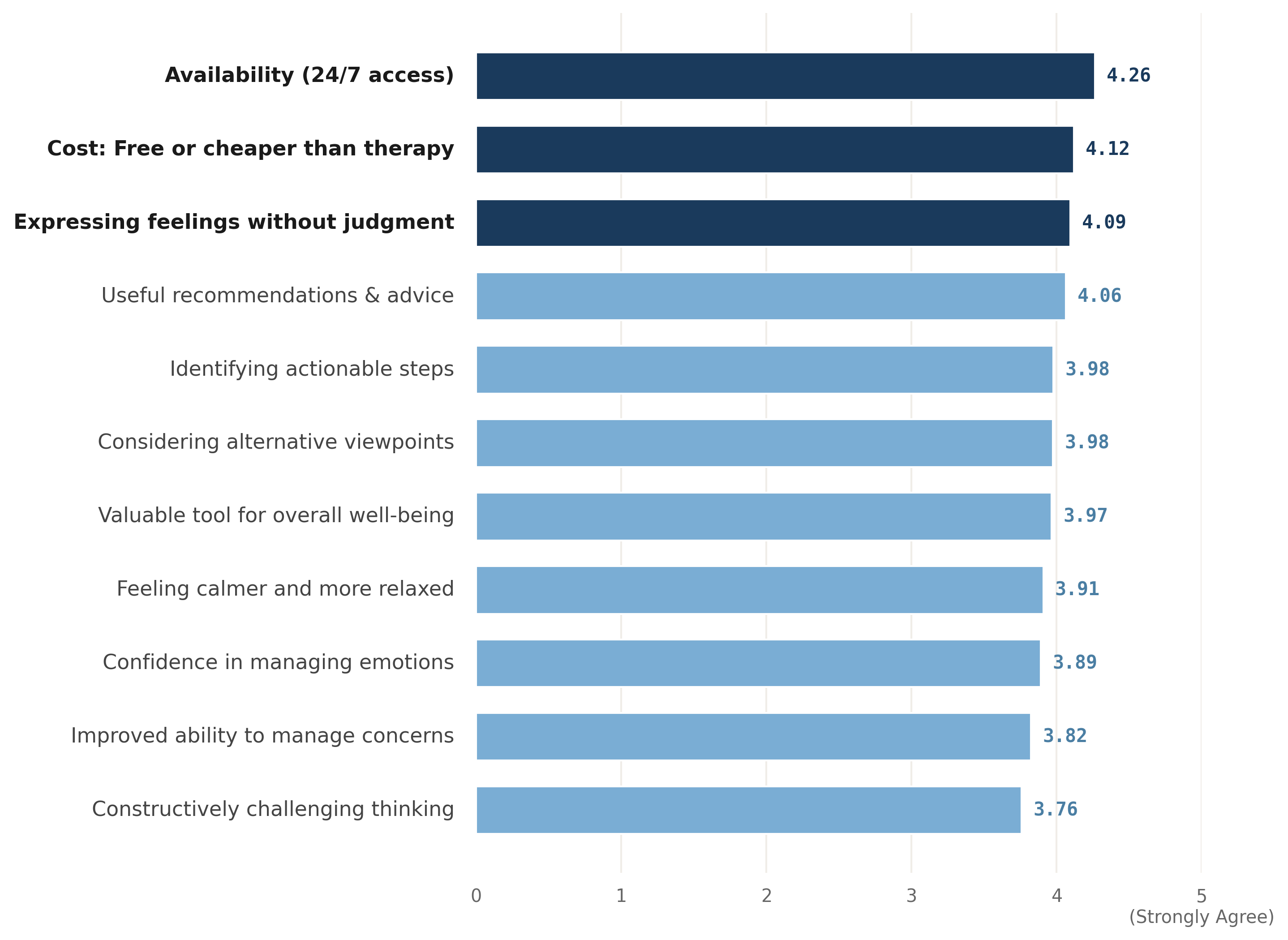}
    \caption{Top-rated Perceived Benefits of AI conversational agents for mental wellbeing and emotional support. Global average scores (Scale: 1-5). Top 3 benefits highlighted.}
    \label{fig:topBenefits}
\end{figure*}

\begin{figure*}[t]
    \centering
    \includegraphics[width=0.8\textwidth]{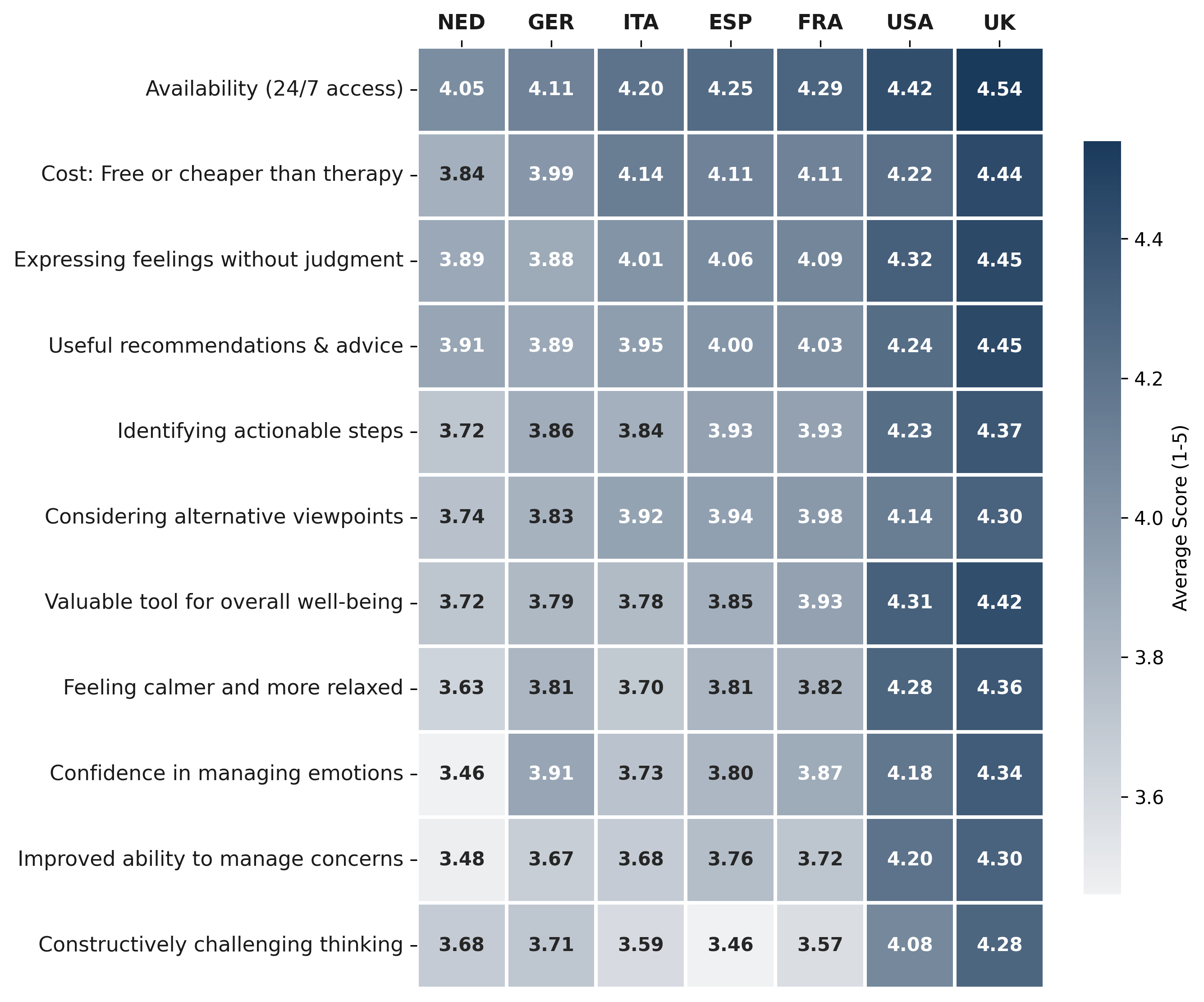}
    \caption{Perceived Benefits across countries. Average scores (Scale: 1-5). Countries ordered by overall enthusiasm (Left to Right).}
    \label{fig:topBenefits_byCountry}
\end{figure*}

Figure \ref{fig:topBenefits} shows the key benefits of seeking support from AI agents, ranked by their average scores provided by users (from 1 to 5). In Figure \ref{fig:topBenefits_byCountry}, we present the average scores for different Perceived Benefits across countries.


\clearpage
\section{User and Non-User Demographic Breakdown}\label{sec:app_demographics}
This section provides the total sample sizes of users and non-users across countries (Table~\ref{tab:ChatbotUsage}), a full demographic breakdown of both user groups, all countries combined (Table~\ref{tab:UserNonUser_Demographics}), and a demographic breakdown of user groups per country (Table~\ref{tab:Demographics_ByCountry}).

\begin{table}[ht]
    \centering
    \small
    \begin{tabular}{lrrrr}
    \toprule
    \textbf{Country} & \textbf{Users} & \textbf{\makecell{Non\\Users}} & \textbf{Total} & \textbf{\% User}  \\
    \midrule
    ESP & 200 & 208 & 408  & 49.0\%\\
    FRA & 190 & 750 & 940  & 20.2\% \\
    GER & 192 & 590 & 782  & 24.6\% \\
    ITA & 198 & 480 & 678  & 29.2\% \\
    NED & 195 & 741 & 936  & 20.8\% \\
    UK  & 184 & 128 & 312  & 59.0\% \\
    USA & 184 & 401 & 585  & 31.5\% \\
    \midrule
    \textbf{Total} & \textbf{1,343} & \textbf{3,298} & \textbf{4,641} & \textbf{28.9\%} \\
    \bottomrule
    \end{tabular}
    \caption{Adoption rates of AI conversational agents for mental wellbeing and emotional support across seven countries.}
    \label{tab:ChatbotUsage}
\end{table}

\begin{table}[ht]
    \centering
    \scriptsize
    \begin{tabular}{@{}lccc@{}}
    \toprule
    \textbf{} & \textbf{Users} & \textbf{Non-Users} & \textbf{Total} \\
     & $1{,}343$ & $3{,}298$ & $4{,}641$ \\
    \midrule
    \textbf{Gender (\%)} & & & \\
    \quad Men & 660 (49.1) & 1{,}598 (48.5) & 2{,}258 (48.7) \\
    \quad Women & \textbf{681 (50.7)} & \textbf{1{,}687 (51.2)} & \textbf{2{,}368 (51.0)} \\
    \quad Non-binary/Other & 2 (0.2) & 13 (0.3) & 15 (0.3) \\
    \addlinespace
    \textbf{Age (\%)} & & & \\
    \quad Under 18 & 2 (0.1) & 22 (0.7) & 24 (0.5) \\
    \quad 18--24 & 109 (8.1) & 78 (2.4) & 187 (4.0) \\
    \quad 25--34 & \textbf{409 (30.5)} & 237 (7.2) & 646 (13.9) \\
    \quad 35--44 & 370 (27.6) & 459 (13.9) & 829 (17.9) \\
    \quad 45--54 & 249 (18.5) & 697 (21.1) & 946 (20.4) \\
    \quad 55--64 & 133 (9.9) & 808 (24.5) & 941 (20.3) \\
    \quad 65+ & 71 (5.3) & \textbf{997 (30.2)} & \textbf{1{,}068 (23.0)} \\
    \addlinespace
    \textbf{Education (\%)} & & & \\
    \quad Primary school & 17 (1.3) & 177 (5.4) & 194 (4.2) \\
    \quad HS & 277 (20.7) & \textbf{1{,}201 (36.8)} & \textbf{1{,}478 (32.1)} \\
    \quad Some college, no degree & 83 (6.2) & 230 (7.1) & 313 (6.8) \\
    \quad Associate degree & 138 (10.3) & 480 (14.7) & 618 (13.4) \\
    \quad BSc & \textbf{308 (23.0)} & 465 (14.3) & 773 (16.8) \\
    \quad MSc & 263 (19.6) & 290 (8.9) & 553 (12.0) \\
    \quad PhD & 78 (5.8) & 60 (1.8) & 138 (3.0) \\
    \quad Professional degree & 175 (13.1) & 357 (11.0) & 532 (11.6) \\
    \addlinespace
    \textbf{Religion (\%)} & & & \\
    \quad Christianity & \textbf{797 (59.3)} & \textbf{1{,}791 (54.3)} & \textbf{2{,}588 (55.8)} \\
    \quad Islam & 83 (6.2) & 78 (2.4) & 161 (3.5) \\
    \quad No religion & 373 (27.8) & 1{,}184 (35.9) & 1{,}557 (33.6) \\
    \quad Other & 52 (3.9) & 119 (3.6) & 171 (3.7) \\
    \quad Prefer not to say & 38 (2.8) & 126 (3.8) & 164 (3.5) \\
    \addlinespace
    \textbf{Marital Status (\%)} & & & \\
    \quad Single & 385 (28.7) & 807 (24.5) & 1{,}192 (25.7) \\
    \quad Married/Partnership & \textbf{864 (64.3)} & \textbf{1{,}881 (57.0)} & \textbf{2{,}745 (59.1)} \\
    \quad Divorced & 62 (4.6) & 342 (10.4) & 404 (8.7) \\
    \quad Separated & 9 (0.7) & 60 (1.8) & 69 (1.5) \\
    \quad Widowed & 15 (1.1) & 179 (5.4) & 194 (4.2) \\
    \quad Prefer not to say & 8 (0.6) & 29 (0.9) & 37 (0.8) \\
    \addlinespace
    \textbf{SES (\%)} & & & \\
    \quad 1 (Worst off) & 11 (0.8) & 66 (2.0) & 77 (1.7) \\
    \quad 2 & 17 (1.3) & 96 (2.9) & 113 (2.4) \\
    \quad 3 & 48 (3.6) & 220 (6.7) & 268 (5.8) \\
    \quad 4 & 114 (8.5) & 340 (10.3) & 454 (9.8) \\
    \quad 5 & 181 (13.5) & 658 (20.0) & 839 (18.1) \\
    \quad 6 & 240 (17.9) & \textbf{717 (21.7)} & 957 (20.6) \\
    \quad 7 & \textbf{306 (22.8)} & 688 (20.9) & \textbf{994 (21.4)} \\
    \quad 8 & 239 (17.8) & 341 (10.3) & 580 (12.5) \\
    \quad 9 & 101 (7.5) & 69 (2.1) & 170 (3.7) \\
    \quad 10 (Best off) & 86 (6.4) & 103 (3.1) & 189 (4.1) \\
    \bottomrule
    \end{tabular}
    \caption{Breakdown of demographic differences between users and non-users. The most common groups are boldfaced.}
    \label{tab:UserNonUser_Demographics}
\end{table}

\begin{sidewaystable*}[p]
    \centering
    \scriptsize
    \setlength{\tabcolsep}{7.5pt}
    \begin{tabular}{@{}lcccccccccccccccccccccc@{}}
    \toprule
     & \multicolumn{3}{c}{\textbf{ESP}} & \multicolumn{3}{c}{\textbf{FRA}} & \multicolumn{3}{c}{\textbf{ITA}} & \multicolumn{3}{c}{\textbf{GER}} & \multicolumn{3}{c}{\textbf{NED}} & \multicolumn{3}{c}{\textbf{UK}} & \multicolumn{3}{c}{\textbf{USA}} \\
    \cmidrule(lr){2-4}\cmidrule(lr){5-7}\cmidrule(lr){8-10}\cmidrule(lr){11-13}\cmidrule(lr){14-16}\cmidrule(lr){17-19}\cmidrule(lr){20-22}
     & U & NU & T & U & NU & T & U & NU & T & U & NU & T & U & NU & T & U & NU & T & U & NU & T \\
     & 200 & 208 & 408 & 190 & 750 & 940 & 198 & 480 & 678 & 192 & 590 & 782 & 195 & 741 & 936 & 184 & 128 & 312 & 184 & 401 & 585 \\
    \midrule
    \textbf{Gender (\%)} & & & & & & & & & & & & & & & & & & & & & \\
    \quad Men
      & 99  & 79  & 178  & 91  & 373  & 464  & \textbf{99 } & \textbf{269 } & \textbf{368 } & 94  & \textbf{314 } & \textbf{408 } & 95  & \textbf{384 } & \textbf{479 } & 90  & \textbf{67 } & \textbf{157 } & \textbf{92 } & 112  & 204  \\
    \quad Women
      & \textbf{100 } & \textbf{127 } & \textbf{227 } & \textbf{98 } & \textbf{377 } & \textbf{475 } & \textbf{99 } & 210  & 309  & \textbf{98 } & 273  & 371  & \textbf{100 } & 354  & 454  & \textbf{94 } & 61  & 155  & \textbf{92 } & \textbf{285 } & \textbf{377 } \\
    \quad Non-binary/Other
      & 1  & 2  & 3  & 1  & 0  & 1  & 0  & 1  & 1  & 0  & 3  & 3  & 0  & 3  & 3  & 0  & 0  & 0  & 0  & 4  & 4  \\
    \addlinespace
    \textbf{Age (\%)} & & & & & & & & & & & & & & & & & & & & & \\
    \quad Under 18
      & 0  & 3  & 3  & 1  & 6  & 7  & 0  & 2  & 2  & 1  & 3  & 4  & 0  & 5  & 5  & 0  & 1  & 1  & 0  & 2  & 2  \\
    \quad 18--24
      & 22  & 5  & 27  & 21  & 27  & 48  & 16  & 7  & 23  & 22  & 14  & 36  & 16  & 11  & 27  & 8  & 0  & 8  & 4  & 14  & 18  \\
    \quad 25--34
      & 53  & 21  & 74  & \textbf{60 } & 53  & 113  & \textbf{55 } & 35  & 90  & 39  & 30  & 69  & 44  & 41  & 85  & \textbf{117 } & 23  & \textbf{140 } & 41  & 34  & 75  \\
    \quad 35--44
      & 42  & 31  & 73  & 45  & 93  & 138  & 43  & 78  & 121  & \textbf{51 } & 77  & 128  & \textbf{58 } & 114  & 172  & 44  & 19  & 63  & \textbf{87 } & 47  & 134  \\
    \quad 45--54
      & \textbf{65 } & \textbf{69 } & \textbf{134 } & 33  & 178  & 211  & 48  & 134  & \textbf{182 } & 28  & 102  & 130  & 38  & 135  & 173  & 10  & 16  & 26  & 27  & 63  & 90  \\
    \quad 55--64
      & 16  & 57  & 73  & 17  & \textbf{202 } & \textbf{219 } & 27  & \textbf{137 } & 164  & 34  & 160  & 194  & 22  & 165  & 187  & 4  & 25  & 29  & 13  & 62  & 75  \\
    \quad 65+
      & 2  & 22  & 24  & 13  & 191  & 204  & 9  & 87  & 96  & 17  & \textbf{204 } & \textbf{221 } & 17  & \textbf{270 } & \textbf{287 } & 1  & \textbf{44 } & 45  & 12  & \textbf{179 } & \textbf{191 } \\
    \addlinespace
    \textbf{Education (\%)} & & & & & & & & & & & & & & & & & & & & & \\
    \quad Primary school
      & 2  & 10  & 12  & 5  & 57  & 62  & 3  & 35  & 38  & 4  & 50  & 54  & 2  & 13  & 15  & 0  & 2  & 2  & 1  & 10  & 11  \\
    \quad HS
      & 18  & 24  & 42  & \textbf{61 } & \textbf{351 } & \textbf{412 } & \textbf{68 } & \textbf{263 } & \textbf{331 } & \textbf{59 } & 155  & 214  & 36  & 236  & 272  & 9  & \textbf{37 } & 46  & 26  & \textbf{135 } & \textbf{161 } \\
    \quad Some college
      & 8  & 9  & 17  & 11  & 41  & 52  & 25  & 36  & 61  & 15  & 23  & 38  & 2  & 13  & 15  & 9  & 23  & 32  & 13  & 85  & 98  \\
    \quad Associate degree
      & 31  & 41  & 72  & 39  & 125  & 164  & 9  & 15  & 24  & 1  & 4  & 5  & \textbf{48 } & \textbf{248 } & \textbf{296 } & 0  & 2  & 2  & 10  & 45  & 55  \\
    \quad BSc
      & 59  & \textbf{52 } & 111  & 35  & 84  & 119  & 38  & 40  & 78  & 30  & 47  & 77  & \textbf{48 } & 126  & 174  & 46  & 33  & \textbf{79 } & 52  & 83  & 135  \\
    \quad MSc
      & 12  & 10  & 22  & 30  & 61  & 91  & 32  & 43  & 75  & 27  & 64  & 91  & 41  & 67  & 108  & \textbf{58 } & 13  & 71  & \textbf{63 } & 32  & 95  \\
    \quad PhD
      & 8  & 10  & 18  & 2  & 11  & 13  & 4  & 7  & 11  & 4  & 6  & 10  & 5  & 10  & 15  & 45  & 10  & 55  & 10  & 6  & 16  \\
    \quad Professional degree
      & \textbf{62 } & 50  & \textbf{112 } & 6  & 10  & 16  & 19  & 39  & 58  & 50  & \textbf{224 } & \textbf{274 } & 13  & 22  & 35  & 16  & 7  & 23  & 9  & 5  & 14  \\
    \addlinespace
    \textbf{Religion (\%)} & & & & & & & & & & & & & & & & & & & & & \\
    \quad Christianity
      & \textbf{124 } & \textbf{129 } & \textbf{253 } & \textbf{87 } & \textbf{368 } & \textbf{455 } & \textbf{142 } & \textbf{367 } & \textbf{509 } & \textbf{100 } & \textbf{286 } & \textbf{386 } & 67  & 287  & 354  & \textbf{153 } & \textbf{83 } & \textbf{236 } & \textbf{124 } & \textbf{271 } & \textbf{395 } \\
    \quad Islam
      & 2  & 1  & 3  & 24  & 35  & 59  & 2  & 2  & 4  & 12  & 13  & 25  & 15  & 19  & 34  & 8  & 1  & 9  & 20  & 7  & 27  \\
    \quad No religion
      & 65  & 68  & 133  & 68  & 294  & 362  & 41  & 91  & 132  & 66  & 260  & 326  & \textbf{100 } & \textbf{370 } & \textbf{470 } & 16  & 36  & 52  & 17  & 65  & 82  \\
    \quad Other
      & 4  & 7  & 11  & 4  & 13  & 17  & 6  & 12  & 18  & 7  & 15  & 22  & 9  & 24  & 33  & 1  & 4  & 5  & 21  & 44  & 65  \\
    \quad Prefer not to say
      & 5  & 3  & 8  & 7  & 40  & 47  & 7  & 8  & 15  & 7  & 16  & 23  & 4  & 41  & 45  & 6  & 4  & 10  & 2  & 14  & 16  \\
    \addlinespace
    \textbf{Marital Status (\%)} & & & & & & & & & & & & & & & & & & & & & \\
    \quad Single
      & 55  & 50  & 105  & 69  & 172  & 241  & 82  & 152  & 234  & 66  & 144  & 210  & 67  & 157  & 224  & 22  & 30  & 52  & 24  & 102  & 126  \\
    \quad Married/Partnership
      & \textbf{130 } & \textbf{123 } & \textbf{253 } & \textbf{114 } & \textbf{467 } & \textbf{581 } & \textbf{102 } & \textbf{263 } & \textbf{365 } & \textbf{103 } & \textbf{320 } & \textbf{423 } & \textbf{110 } & \textbf{447 } & \textbf{557 } & \textbf{160 } & \textbf{82 } & \textbf{242 } & \textbf{145 } & \textbf{179 } & \textbf{324 } \\
    \quad Divorced
      & 11  & 26  & 37  & 6  & 61  & 67  & 6  & 37  & 43  & 11  & 79  & 90  & 16  & 86  & 102  & 2  & 8  & 10  & 10  & 45  & 55  \\
    \quad Separated
      & 2  & 3  & 5  & 1  & 21  & 22  & 0  & 9  & 9  & 3  & 11  & 14  & 1  & 5  & 6  & 0  & 2  & 2  & 2  & 9  & 11  \\
    \quad Widowed
      & 2  & 4  & 6  & 0  & 26  & 26  & 2  & 14  & 16  & 7  & 32  & 39  & 1  & 36  & 37  & 0  & 5  & 5  & 3  & 62  & 65  \\
    \quad Prefer not to say
      & 0  & 2  & 2  & 0  & 3  & 3  & 6  & 5  & 11  & 2  & 4  & 6  & 0  & 10  & 10  & 0  & 1  & 1  & 0  & 4  & 4  \\
    \addlinespace
    \textbf{SES (\%)} & & & & & & & & & & & & & & & & & & & & & \\
    \quad 1 (Worst off)
      & 1  & 1  & 2  & 1  & 13  & 14  & 1  & 10  & 11  & 3  & 17  & 20  & 2  & 6  & 8  & 1  & 1  & 2  & 2  & 18  & 20  \\
    \quad 2
      & 1  & 1  & 2  & 6  & 28  & 34  & 3  & 3  & 6  & 2  & 21  & 23  & 3  & 21  & 24  & 1  & 4  & 5  & 1  & 18  & 19  \\
    \quad 3
      & 5  & 10  & 15  & 16  & 51  & 67  & 4  & 30  & 34  & 12  & 45  & 57  & 3  & 31  & 34  & 1  & 9  & 10  & 7  & 44  & 51  \\
    \quad 4
      & 22  & 12  & 34  & 31  & 114  & 145  & 22  & 45  & 67  & 12  & 76  & 88  & 11  & 44  & 55  & 5  & 15  & 20  & 11  & 34  & 45  \\
    \quad 5
      & 30  & 44  & 74  & 35  & \textbf{205 } & \textbf{240 } & 39  & 98  & 137  & 28  & \textbf{126 } & 154  & 16  & 81  & 97  & 12  & \textbf{32 } & 44  & 21  & \textbf{72 } & 93  \\
    \quad 6
      & \textbf{53 } & \textbf{57 } & \textbf{110 } & \textbf{46 } & 166  & 212  & \textbf{44 } & \textbf{144 } & \textbf{188 } & \textbf{39 } & 118  & \textbf{157 } & 23  & 146  & 169  & 13  & 22  & 35  & 22  & 64  & 86  \\
    \quad 7
      & 46  & 45  & 91  & 32  & 105  & 137  & 40  & 98  & 138  & \textbf{39 } & 105  & 144  & \textbf{71 } & \textbf{245 } & \textbf{316 } & \textbf{45 } & 21  & \textbf{66 } & 33  & 69  & \textbf{102 } \\
    \quad 8
      & 26  & 24  & 50  & 10  & 48  & 58  & 33  & 29  & 62  & 34  & 55  & 89  & 54  & 136  & 190  & 44  & 12  & 56  & \textbf{38 } & 37  & 75  \\
    \quad 9
      & 12  & 6  & 18  & 5  & 6  & 11  & 6  & 9  & 15  & 14  & 12  & 26  & 8  & 19  & 27  & 29  & 3  & 32  & 27  & 14  & 41  \\
    \quad 10 (Best off)
      & 4  & 8  & 12  & 8  & 14  & 22  & 6  & 14  & 20  & 9  & 15  & 24  & 4  & 12  & 16  & 33  & 9  & 42  & 22  & 31  & 53  \\
    \bottomrule
    \end{tabular}
    \caption{Demographic breakdown by country for users (U), non-users (NU), and total (T). The most common group within each demographic category per country is boldfaced.}
    \label{tab:Demographics_ByCountry}
\end{sidewaystable*}

\clearpage
\section{Geographic Differences in Trust \& Privacy Perceptions}\label{app:appc}

\begin{figure*}[htbp]
    \centering
    \begin{subfigure}[t]{\textwidth}
        \centering
        \includegraphics[width=\textwidth]{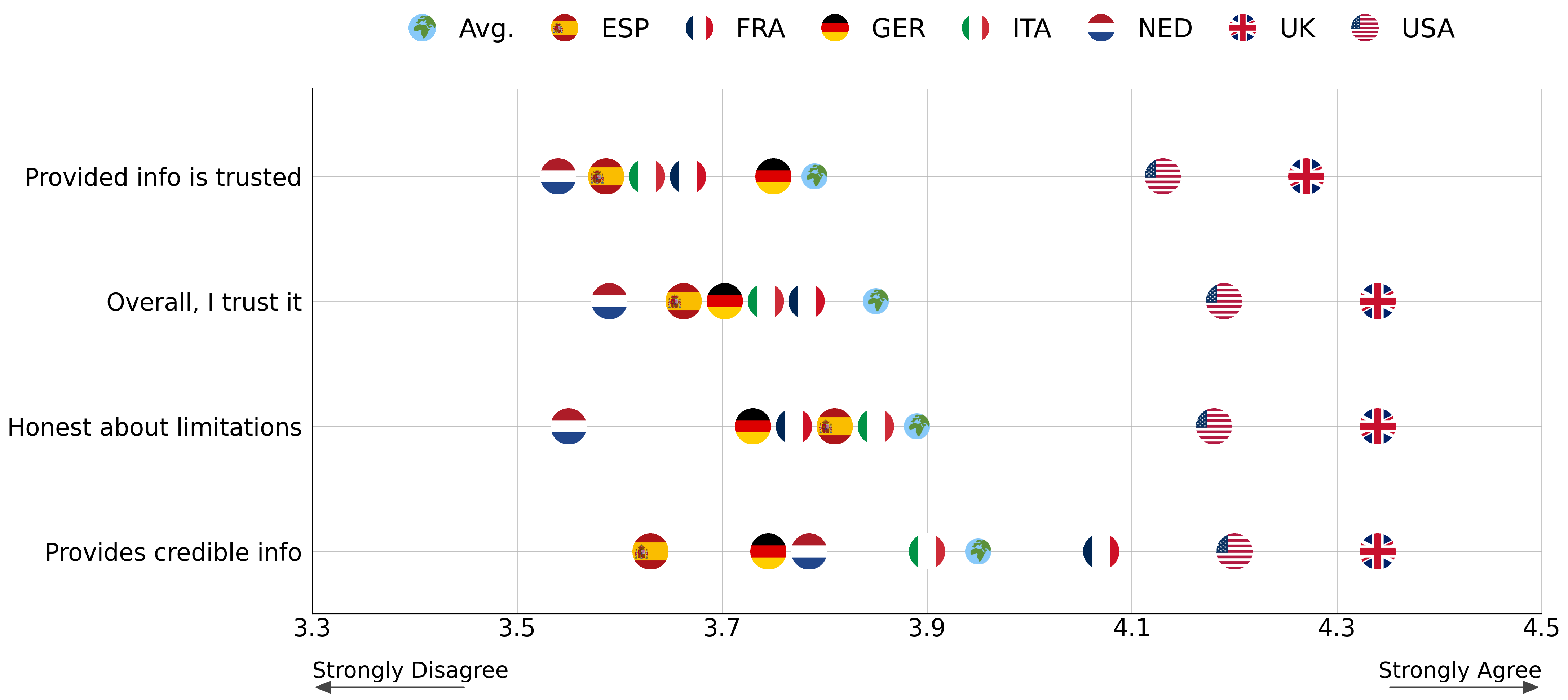}
        \caption{Trust perceptions}
        \label{fig:trust-perceptions}
    \end{subfigure}
    \vspace{0.3cm}
    \begin{subfigure}[t]{\textwidth}
        \centering
        \includegraphics[width=\textwidth]{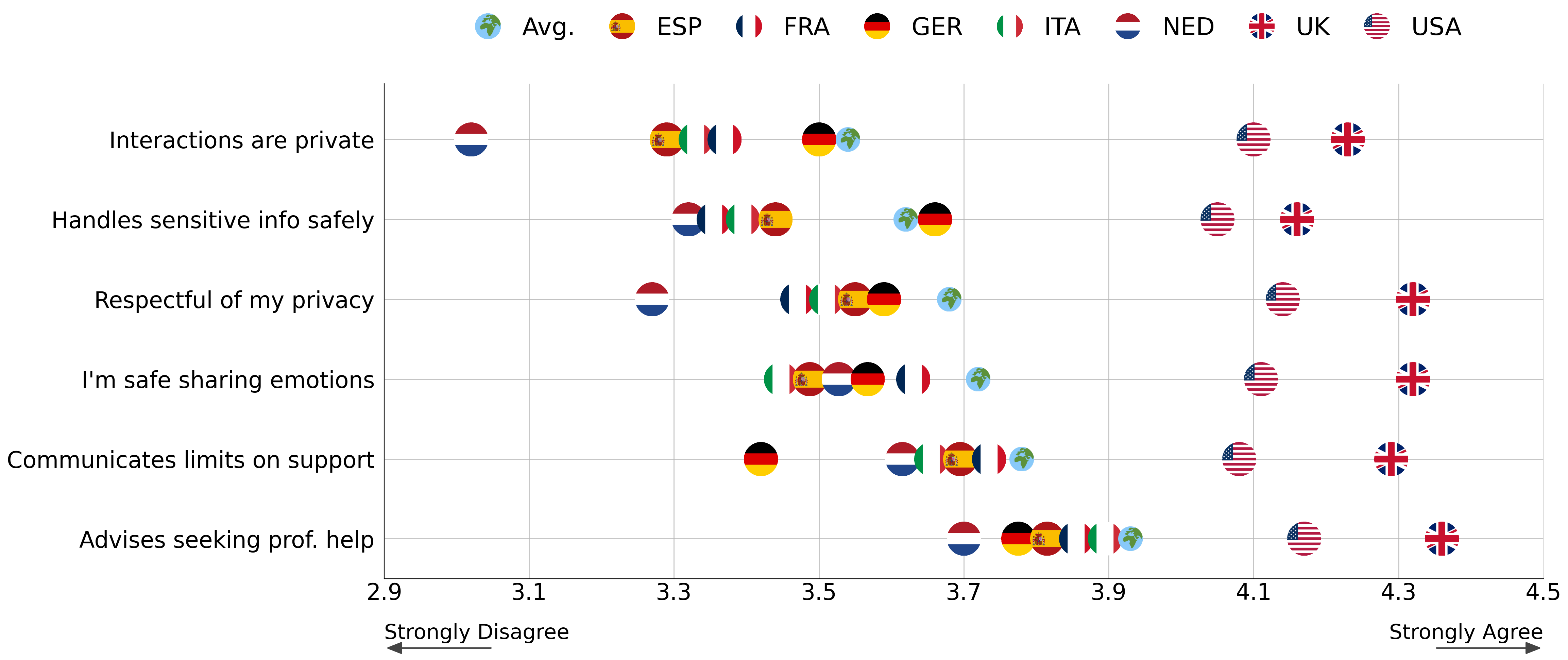}
        \caption{Privacy perceptions}
        \label{fig:privacy-perceptions}
    \end{subfigure}
    \caption{Trust and privacy perceptions by country. Items sorted by global average (ascending). Scores represent mean agreement on a 1--5 Likert scale.}
    \label{fig:trust-privacy-perceptions}
\end{figure*}

Figure \ref{fig:trust-privacy-perceptions} presents mean agreement scores for trust and privacy perceptions of mental wellbeing chatbots across seven countries, with items sorted by ascending global average. In both panels, the UK and USA consistently score highest, clustering toward the 'Strongly Agree' end of the scale, while continental European countries tend to rate lower. Trust perceptions show relatively tight cross-country clustering on items like 'Provided info is trusted,' but greater spread on 'Provides credible info' and 'Honest about limitations.' Privacy perceptions exhibit a similar UK/USA–continental Europe divide, with the largest cross-country variation appearing on items related to emotional safety and professional help advice. Across both constructs, the Netherlands usually has the most negative perceptions, distinguishing itself from other European countries. Dutch participants especially rate LLMs' honesty about limitations and interaction privacy low, showing skepticism.

\clearpage
\section{Descriptive Statistics for Constructs}\label{app:app_b}
\label{DescStatsConstructs}

\begin{figure}[t]
    \centering
    \includegraphics[width=\columnwidth]{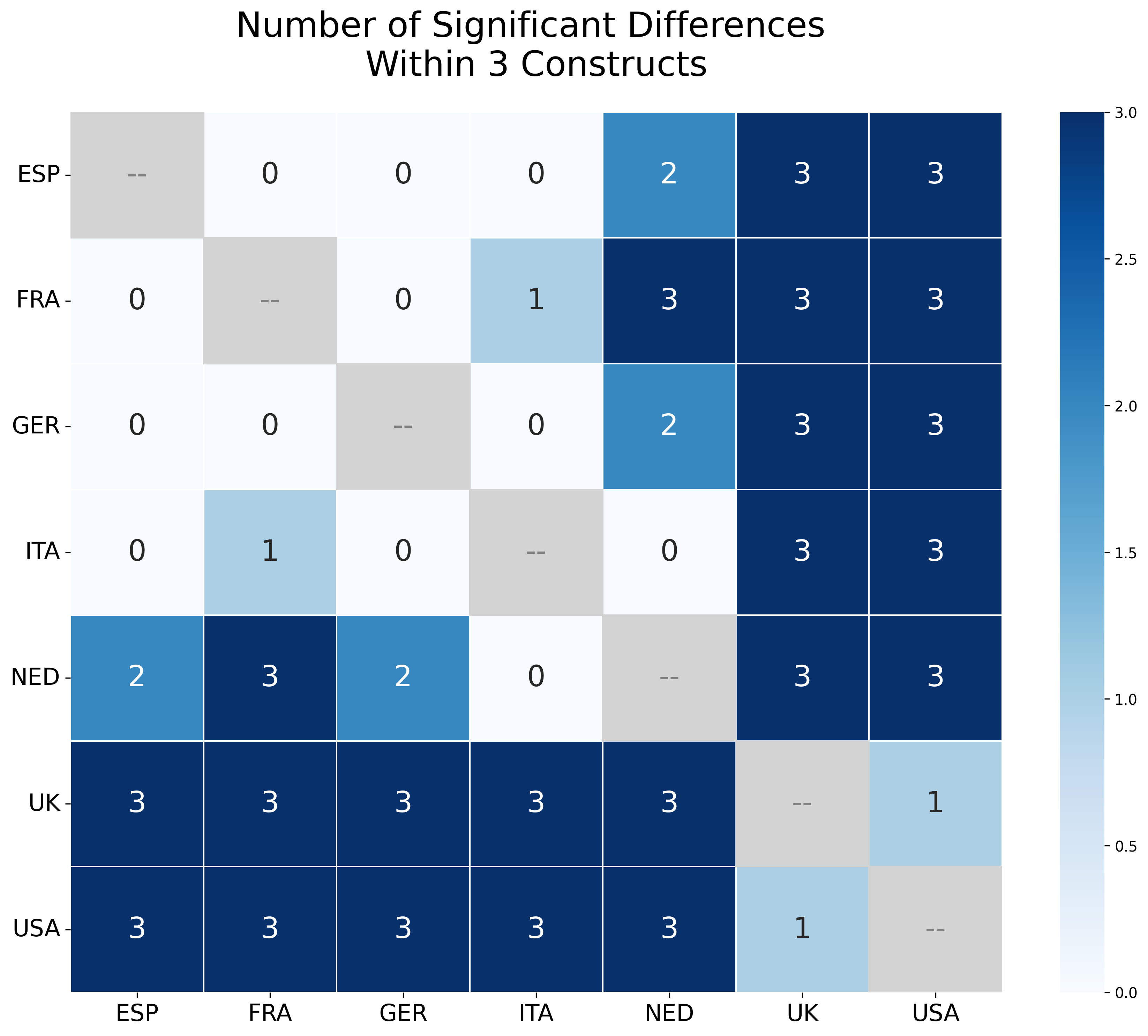}
    \caption{Pairwise country comparisons showing the number of constructs (out of 3) that differ significantly between each pair. The UK and the USA clearly deviate from the mainland European countries and form their own cluster.}
    \label{fig:PairwiseCountryComparison}
\end{figure}

Descriptive statistics for all constructs per country are shown in Table~\ref{tab:DescriptiveByCountry}. The demographic breakdown of constructs is represented in Table~\ref{tab:DescriptiveByDemographic}. The full statistical test results mentioned in \S\ref{culture_constructs} are displayed in Table~\ref{tab:PairwiseCountryConstructs}, and visualized in Figure~\ref{fig:PairwiseCountryComparison}. 

Looking at the tables, the UK has the highest mean and median scores in all constructs, with the US following closely. The table and the figure confirm that both the UK and the USA statistically differ from the rest of the countries across all constructs. Within, they only differ in BENF. It is evident that the Anglosphere forms a separate cluster. Following this, the remaining countries form their own cluster: Germany does not statistically differ from any mainland European country except the Netherlands. Italy aligns with the Netherlands, Spain, and Germany in all constructs and differs from France in one construct. Spain differs from the Netherlands in two constructs. This pattern confirms that the Netherlands distinguishes itself from the others.

\begin{table}[t]
    \fontsize{6.5pt}{5.0pt}\selectfont
    \renewcommand{\arraystretch}{1.3}
    \setlength{\tabcolsep}{5pt}
    \begin{tabular}{@{}l*{3}{ccc}@{}}
    \toprule
    & \multicolumn{3}{c}{\textbf{BENF}} & \multicolumn{3}{c}{\textbf{USE}} & \multicolumn{3}{c}{\textbf{TRU}} \\
    \cmidrule(lr){2-4} \cmidrule(lr){5-7} \cmidrule(lr){8-10}
    \textbf{Country} & $\bar{x}$ & $\tilde{x}$ & $\sigma$ & $\bar{x}$ & $\tilde{x}$ & $\sigma$ & $\bar{x}$ & $\tilde{x}$ & $\sigma$ \\
    \midrule
    Combined
      & 4.00 & 4.00 & 0.62
      & 3.77 & 3.86 & 0.80
      & 3.78 & 3.80 & 0.72 \\
    ESP
      & 3.91 & 4.00 & 0.69
      & 3.64 & 3.71 & 0.85
      & 3.60 & 3.70 & 0.84 \\
    FRA
      & 3.95 & 4.00 & 0.64
      & 3.69 & 3.86 & 0.85
      & 3.67 & 3.80 & 0.74 \\
    ITA
      & 3.89 & 3.90 & 0.55
      & 3.53 & 3.57 & 0.73
      & 3.63 & 3.70 & 0.62 \\
    GER
      & 3.87 & 4.00 & 0.67
      & 3.68 & 3.86 & 0.81
      & 3.65 & 3.80 & 0.71 \\
    NED
      & 3.75 & 3.80 & 0.56
      & 3.42 & 3.43 & 0.79
      & 3.49 & 3.60 & 0.59 \\
    UK
      & \textbf{4.39} & \textbf{4.40} & 0.44
      & \textbf{4.22} & \textbf{4.29} & 0.61
      & \textbf{4.30} & \textbf{4.40} & 0.53 \\
    USA
      & 4.24 & 4.30 & 0.46
      & 4.17 & 4.21 & 0.53
      & 4.14 & 4.25 & 0.59 \\
    \bottomrule
    \end{tabular}
    \caption{Descriptive statistics for each construct by country. $\bar{x}$ = mean; $\tilde{x}$ = median; $\sigma$ = standard deviation. The highest mean and median scores are bold-faced.}
    \label{tab:DescriptiveByCountry}
\end{table}

\begin{table*}[t]
    \centering
    \small
    \begin{tabular}{@{}l*{3}{ccc}@{}}
    \toprule
    & \multicolumn{3}{c}{\textbf{BENF}} & \multicolumn{3}{c}{\textbf{USE}} & \multicolumn{3}{c}{\textbf{TRU}} \\
    \cmidrule(lr){2-4} \cmidrule(lr){5-7} \cmidrule(lr){8-10}
    & $\bar{x}$ & $\tilde{x}$ & $\sigma$ & $\bar{x}$ & $\tilde{x}$ & $\sigma$ & $\bar{x}$ & $\tilde{x}$ & $\sigma$ \\
    \midrule
    \textbf{Gender} & & & & & & & & & \\
    \quad Female
      & \textbf{4.00} & \textbf{4.00} & 0.61
      & 3.73 & \textbf{3.86} & 0.82
      & 3.74 & 3.80 & 0.74 \\
    \quad Male
      & 3.99 & \textbf{4.00} & 0.62
      & \textbf{3.79} & \textbf{3.86} & 0.77
      & \textbf{3.81} & \textbf{3.90} & 0.71 \\
    \addlinespace
    \textbf{Age} & & & & & & & & & \\
    \quad 18--24
      & 3.89 & 4.00 & 0.65
      & 3.57 & 3.71 & 0.85
      & 3.55 & 3.70 & 0.81 \\
    \quad 25--34
      & \textbf{4.07} & \textbf{4.10} & 0.60
      & \textbf{3.87} & \textbf{4.00} & 0.76
      & \textbf{3.87} & \textbf{3.90} & 0.70 \\
    \quad 35--44
      & 4.03 & \textbf{4.10} & 0.59
      & 3.85 & \textbf{4.00} & 0.75
      & 3.86 & \textbf{3.90} & 0.70 \\
    \quad 45--54
      & 3.97 & 4.00 & 0.63
      & 3.67 & 3.86 & 0.81
      & 3.69 & 3.80 & 0.77 \\
    \quad 55--64
      & 3.88 & 4.00 & 0.63
      & 3.61 & 3.71 & 0.84
      & 3.67 & 3.80 & 0.67 \\
    \quad 65+
      & 3.82 & 3.80 & 0.64
      & 3.48 & 3.57 & 0.87
      & 3.59 & 3.70 & 0.61 \\
    \addlinespace
    \textbf{Education} & & & & & & & & & \\
    \quad Primary school
      & 4.04 & 3.90 & 0.62
      & 3.78 & 4.00 & 0.91
      & 3.67 & 3.50 & 0.64 \\
    \quad High school
      & 3.91 & 4.00 & 0.62
      & 3.67 & 3.71 & 0.79
      & 3.71 & 3.80 & 0.66 \\
    \quad Some college
      & 3.98 & 4.00 & 0.64
      & 3.58 & 3.71 & 0.86
      & 3.70 & 3.80 & 0.73 \\
    \quad Associate degree
      & 3.85 & 3.85 & 0.64
      & 3.57 & 3.71 & 0.87
      & 3.58 & 3.70 & 0.81 \\
    \quad BSc
      & 4.03 & 4.10 & 0.57
      & 3.80 & 3.86 & 0.76
      & 3.80 & 3.90 & 0.73 \\
    \quad MSc
      & 4.06 & 4.10 & 0.62
      & 3.82 & 4.00 & 0.82
      & 3.88 & 3.90 & 0.75 \\
    \quad PhD
      & \textbf{4.29} & \textbf{4.40} & 0.51
      & \textbf{4.17} & \textbf{4.29} & 0.63
      & \textbf{4.13} & \textbf{4.20} & 0.68 \\
    \quad Professional degree
      & 3.93 & 4.00 & 0.64
      & 3.76 & 3.86 & 0.74
      & 3.71 & 3.80 & 0.67 \\
    \addlinespace
    \textbf{Religion} & & & & & & & & & \\
    \quad Christianity
      & 4.06 & \textbf{4.10} & 0.60
      & 3.85 & \textbf{4.00} & 0.78
      & 3.87 & 3.90 & 0.72 \\
    \quad Islam
      & \textbf{4.11} & \textbf{4.10} & 0.50
      & \textbf{3.99} & \textbf{4.00} & 0.70
      & \textbf{3.94} & \textbf{4.00} & 0.68 \\
    \quad No religion
      & 3.85 & 3.90 & 0.65
      & 3.50 & 3.57 & 0.83
      & 3.54 & 3.60 & 0.71 \\
    \quad Other
      & 4.00 & \textbf{4.10} & 0.55
      & 3.82 & 3.93 & 0.78
      & 3.83 & 3.90 & 0.70 \\
    \quad Prefer not to say
      & 3.85 & 3.95 & 0.58
      & 3.67 & 3.71 & 0.65
      & 3.65 & 3.70 & 0.66 \\
    \addlinespace
    \textbf{Marital Status} & & & & & & & & & \\
    \quad Single
      & 3.86 & 3.90 & 0.62
      & 3.57 & 3.71 & 0.84
      & 3.57 & 3.60 & 0.72 \\
    \quad Married/Partnership
      & \textbf{4.06} & \textbf{4.10} & 0.60
      & \textbf{3.86} & \textbf{4.00} & 0.77
      & \textbf{3.89} & \textbf{4.00} & 0.71 \\
    \quad Divorced
      & 3.99 & 4.05 & 0.54
      & 3.68 & 3.86 & 0.72
      & 3.66 & 3.80 & 0.51 \\
    \quad Separated
      & 3.34 & 3.60 & 0.57
      & 3.29 & 3.43 & 0.74
      & 3.00 & 3.60 & 0.90 \\
    \quad Widowed
      & 3.90 & 3.90 & 0.64
      & 3.83 & 3.86 & 0.67
      & 3.61 & 3.80 & 0.63 \\
    \quad Prefer not to say
      & 3.51 & 3.35 & 0.62
      & 2.98 & 3.07 & 1.02
      & 3.11 & 3.10 & 0.36 \\
    \addlinespace
    \textbf{SES} & & & & & & & & & \\
    \quad 1 (Worst off)
      & 4.12 & 4.30 & 0.72
      & 3.78 & 4.00 & 1.05
      & 3.93 & 4.00 & 0.79 \\
    \quad 2
      & 3.74 & 3.90 & 0.85
      & 3.59 & 4.00 & 0.96
      & 3.49 & 3.70 & 0.94 \\
    \quad 3
      & 3.90 & 3.90 & 0.71
      & 3.66 & 3.71 & 0.88
      & 3.54 & 3.60 & 0.78 \\
    \quad 4
      & 3.92 & 4.00 & 0.66
      & 3.66 & 3.71 & 0.83
      & 3.61 & 3.65 & 0.75 \\
    \quad 5
      & 3.87 & 3.90 & 0.65
      & 3.54 & 3.57 & 0.79
      & 3.60 & 3.70 & 0.77 \\
    \quad 6
      & 3.99 & 4.00 & 0.57
      & 3.70 & 3.71 & 0.76
      & 3.72 & 3.80 & 0.63 \\
    \quad 7
      & 3.94 & 4.00 & 0.57
      & 3.66 & 3.86 & 0.79
      & 3.69 & 3.75 & 0.69 \\
    \quad 8
      & 3.97 & 4.00 & 0.62
      & 3.79 & 3.86 & 0.78
      & 3.85 & 3.90 & 0.68 \\
    \quad 9
      & 4.28 & 4.30 & 0.41
      & 4.16 & 4.14 & 0.56
      & 4.15 & 4.20 & 0.50 \\
    \quad 10 (Best off)
      & \textbf{4.36} & \textbf{4.50} & 0.62
      & \textbf{4.39} & \textbf{4.57} & 0.63
      & \textbf{4.32} & \textbf{4.60} & 0.78 \\
    \bottomrule
    \end{tabular}
    \caption{Descriptive statistics for each construct by demographic group. $\bar{x}$ = mean; $\tilde{x}$ = median; $\sigma$ = standard deviation. The highest mean and median within each demographic category are bold-faced. Groups with $n < 5$ are omitted.}
    \label{tab:DescriptiveByDemographic}
\end{table*}

\begin{table}[t]
    \scriptsize
    \begin{tabular}{l|rrrrrr}
    \toprule
     & \multicolumn{2}{c}{BENF} & \multicolumn{2}{c}{USE} & \multicolumn{2}{c}{TRU} \\
    \cmidrule(lr){2-3} \cmidrule(lr){4-5} \cmidrule(lr){6-7}
     & \textit{U} & $p_{\text{adj}}$ & \textit{U} & $p_{\text{adj}}$ & \textit{U} & $p_{\text{adj}}$ \\
    \midrule
    UK--USA & \textbf{20035} & \textbf{.021} & 18304 & .878 & 19644 & .069 \\
    UK--ESP & \textbf{26294} & \textbf{$\mathrel{\scalebox{0.7}{<}}$ .001} & \textbf{26142} & \textbf{$\mathrel{\scalebox{0.7}{<}}$ .001} & \textbf{27886} & \textbf{$\mathrel{\scalebox{0.7}{<}}$ .001} \\
    UK--FRA & \textbf{25017} & \textbf{$\mathrel{\scalebox{0.7}{<}}$ .001} & \textbf{24341} & \textbf{$\mathrel{\scalebox{0.7}{<}}$ .001} & \textbf{26482} & \textbf{$\mathrel{\scalebox{0.7}{<}}$ .001} \\
    UK--GER & \textbf{26482} & \textbf{$\mathrel{\scalebox{0.7}{<}}$ .001} & \textbf{25086} & \textbf{$\mathrel{\scalebox{0.7}{<}}$ .001} & \textbf{27288} & \textbf{$\mathrel{\scalebox{0.7}{<}}$ .001} \\
    UK--ITA & \textbf{28062} & \textbf{$\mathrel{\scalebox{0.7}{<}}$ .001} & \textbf{28246} & \textbf{$\mathrel{\scalebox{0.7}{<}}$ .001} & \textbf{29109} & \textbf{$\mathrel{\scalebox{0.7}{<}}$ .001} \\
    UK--NED & \textbf{29501} & \textbf{$\mathrel{\scalebox{0.7}{<}}$ .001} & \textbf{28706} & \textbf{$\mathrel{\scalebox{0.7}{<}}$ .001} & \textbf{30539} & \textbf{$\mathrel{\scalebox{0.7}{<}}$ .001} \\
    \midrule
    USA--UK & \textbf{20035} & \textbf{.021} & 18304 & .878 & 19644 & .069 \\
    USA--ESP & \textbf{23560} & \textbf{$\mathrel{\scalebox{0.7}{<}}$ .001} & \textbf{25414} & \textbf{$\mathrel{\scalebox{0.7}{<}}$ .001} & \textbf{25780} & \textbf{$\mathrel{\scalebox{0.7}{<}}$ .001} \\
    USA--FRA & \textbf{22198} & \textbf{$\mathrel{\scalebox{0.7}{<}}$ .001} & \textbf{23558} & \textbf{$\mathrel{\scalebox{0.7}{<}}$ .001} & \textbf{24339} & \textbf{$\mathrel{\scalebox{0.7}{<}}$ .001} \\
    USA--GER & \textbf{23852} & \textbf{$\mathrel{\scalebox{0.7}{<}}$ .001} & \textbf{24370} & \textbf{$\mathrel{\scalebox{0.7}{<}}$ .001} & \textbf{25166} & \textbf{$\mathrel{\scalebox{0.7}{<}}$ .001} \\
    USA--ITA & \textbf{25302} & \textbf{$\mathrel{\scalebox{0.7}{<}}$ .001} & \textbf{27782} & \textbf{$\mathrel{\scalebox{0.7}{<}}$ .001} & \textbf{27068} & \textbf{$\mathrel{\scalebox{0.7}{<}}$ .001} \\
    USA--NED & \textbf{27024} & \textbf{$\mathrel{\scalebox{0.7}{<}}$ .001} & \textbf{28290} & \textbf{$\mathrel{\scalebox{0.7}{<}}$ .001} & \textbf{28636} & \textbf{$\mathrel{\scalebox{0.7}{<}}$ .001} \\
    \midrule
    ESP--UK & \textbf{26294} & \textbf{$\mathrel{\scalebox{0.7}{<}}$ .001} & \textbf{26142} & \textbf{$\mathrel{\scalebox{0.7}{<}}$ .001} & \textbf{27886} & \textbf{$\mathrel{\scalebox{0.7}{<}}$ .001} \\
    ESP--USA & \textbf{23560} & \textbf{$\mathrel{\scalebox{0.7}{<}}$ .001} & \textbf{25414} & \textbf{$\mathrel{\scalebox{0.7}{<}}$ .001} & \textbf{25780} & \textbf{$\mathrel{\scalebox{0.7}{<}}$ .001} \\
    ESP--FRA & 18415 & 1.000 & 18276 & 1.000 & 18501 & 1.000 \\
    ESP--GER & 20028 & 1.000 & 18797 & 1.000 & 19050 & 1.000 \\
    ESP--ITA & 21062 & 1.000 & 22104 & .266 & 20434 & 1.000 \\
    ESP--NED & \textbf{22992} & \textbf{.021} & \textbf{22960} & \textbf{.020} & 22245 & .123 \\
    \midrule
    FRA--UK & \textbf{25017} & \textbf{$\mathrel{\scalebox{0.7}{<}}$ .001} & \textbf{24341} & \textbf{$\mathrel{\scalebox{0.7}{<}}$ .001} & \textbf{26482} & \textbf{$\mathrel{\scalebox{0.7}{<}}$ .001} \\
    FRA--USA & \textbf{22198} & \textbf{$\mathrel{\scalebox{0.7}{<}}$ .001} & \textbf{23558} & \textbf{$\mathrel{\scalebox{0.7}{<}}$ .001} & \textbf{24339} & \textbf{$\mathrel{\scalebox{0.7}{<}}$ .001} \\
    FRA--ESP & 18415 & 1.000 & 18276 & 1.000 & 18501 & 1.000 \\
    FRA--GER & 16734 & .812 & 17868 & 1.000 & 17962 & 1.000 \\
    FRA--ITA & 20878 & .364 & \textbf{21834} & \textbf{.049} & 19950 & 1.000 \\
    FRA--NED & \textbf{14247} & \textbf{$\mathrel{\scalebox{0.7}{<}}$ .001} & \textbf{14464} & \textbf{.002} & \textbf{15278} & \textbf{.032} \\
    \midrule
    GER--UK & \textbf{26482} & \textbf{$\mathrel{\scalebox{0.7}{<}}$ .001} & \textbf{25086} & \textbf{$\mathrel{\scalebox{0.7}{<}}$ .001} & \textbf{27288} & \textbf{$\mathrel{\scalebox{0.7}{<}}$ .001} \\
    GER--USA & \textbf{23852} & \textbf{$\mathrel{\scalebox{0.7}{<}}$ .001} & \textbf{24370} & \textbf{$\mathrel{\scalebox{0.7}{<}}$ .001} & \textbf{25166} & \textbf{$\mathrel{\scalebox{0.7}{<}}$ .001} \\
    GER--ESP & 20028 & 1.000 & 18797 & 1.000 & 19050 & 1.000 \\
    GER--FRA & 16734 & .812 & 17868 & 1.000 & 17962 & 1.000 \\
    GER--ITA & 19466 & 1.000 & 21743 & .097 & 19988 & 1.000 \\
    GER--NED & 16000 & .106 & \textbf{14962} & \textbf{.006} & \textbf{15535} & \textbf{.037} \\
    \midrule
    ITA--UK & \textbf{28062} & \textbf{$\mathrel{\scalebox{0.7}{<}}$ .001} & \textbf{28246} & \textbf{$\mathrel{\scalebox{0.7}{<}}$ .001} & \textbf{29109} & \textbf{$\mathrel{\scalebox{0.7}{<}}$ .001} \\
    ITA--USA & \textbf{25302} & \textbf{$\mathrel{\scalebox{0.7}{<}}$ .001} & \textbf{27782} & \textbf{$\mathrel{\scalebox{0.7}{<}}$ .001} & \textbf{27068} & \textbf{$\mathrel{\scalebox{0.7}{<}}$ .001} \\
    ITA--ESP & 21062 & 1.000 & 22104 & .266 & 20434 & 1.000 \\
    ITA--FRA & 20878 & .364 & \textbf{21834} & \textbf{.049} & 19950 & 1.000 \\
    ITA--GER & 19466 & 1.000 & 21743 & .097 & 19988 & 1.000 \\
    ITA--NED & 16832 & .194 & 18012 & 1.000 & 16968 & .263 \\
    \midrule
    NED--UK & \textbf{29501} & \textbf{$\mathrel{\scalebox{0.7}{<}}$ .001} & \textbf{28706} & \textbf{$\mathrel{\scalebox{0.7}{<}}$ .001} & \textbf{30539} & \textbf{$\mathrel{\scalebox{0.7}{<}}$ .001} \\
    NED--USA & \textbf{27024} & \textbf{$\mathrel{\scalebox{0.7}{<}}$ .001} & \textbf{28290} & \textbf{$\mathrel{\scalebox{0.7}{<}}$ .001} & \textbf{28636} & \textbf{$\mathrel{\scalebox{0.7}{<}}$ .001} \\
    NED--ESP & \textbf{22992} & \textbf{.021} & \textbf{22960} & \textbf{.020} & 22245 & .123 \\
    NED--FRA & \textbf{14247} & \textbf{$\mathrel{\scalebox{0.7}{<}}$ .001} & \textbf{14464} & \textbf{.002} & \textbf{15278} & \textbf{.032} \\
    NED--GER & 16000 & .106 & \textbf{14962} & \textbf{.006} & \textbf{15535} & \textbf{.037} \\
    NED--ITA & 16832 & .194 & 18012 & 1.000 & 16968 & .263 \\
    \bottomrule
    \end{tabular}
    \caption{Pairwise Mann--Whitney \textit{U} Statistics and Holm-Adjusted \textit{p}-Values Across Constructs. Bold values indicate statistical significance after Holm correction ($p_{\text{adj}} < .05$).}
    \label{tab:PairwiseCountryConstructs}
\end{table}

\clearpage
\section{Mixed Model Analysis of Perceived Benefits, Trust, and Usage}
\label{App_MixedModelConstructs}

\begin{figure}[t]
    \centering
    \includegraphics[width=\columnwidth]{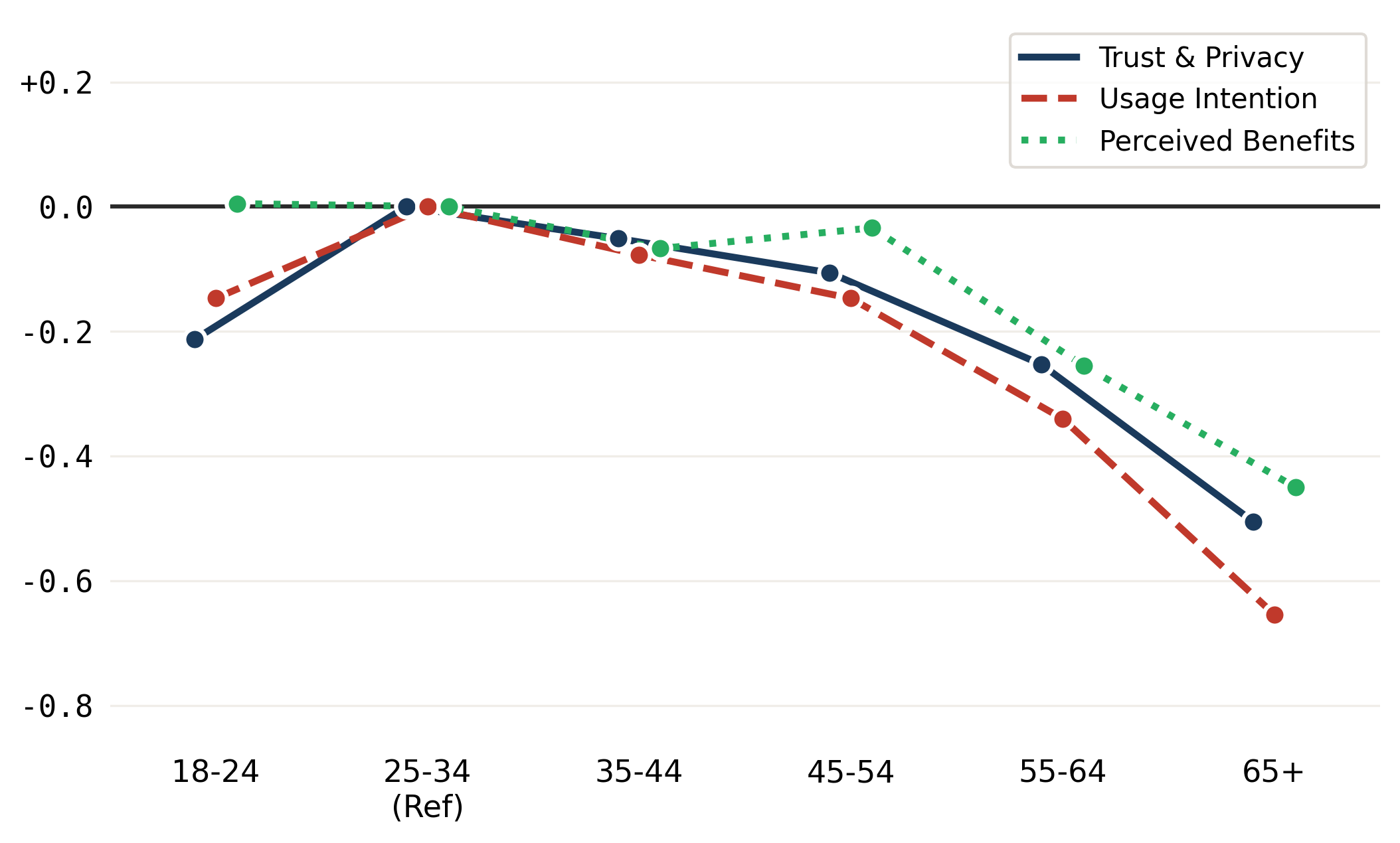}
    \caption{Older generations show lower Trust, Perceived Benefits, and Usage Intention. Fixed effect estimates by age group (Ref = 25-34 years).}
    \label{fig:age_effects}
\end{figure}

\begin{figure}[t]
    \centering
    \includegraphics[width=\columnwidth]{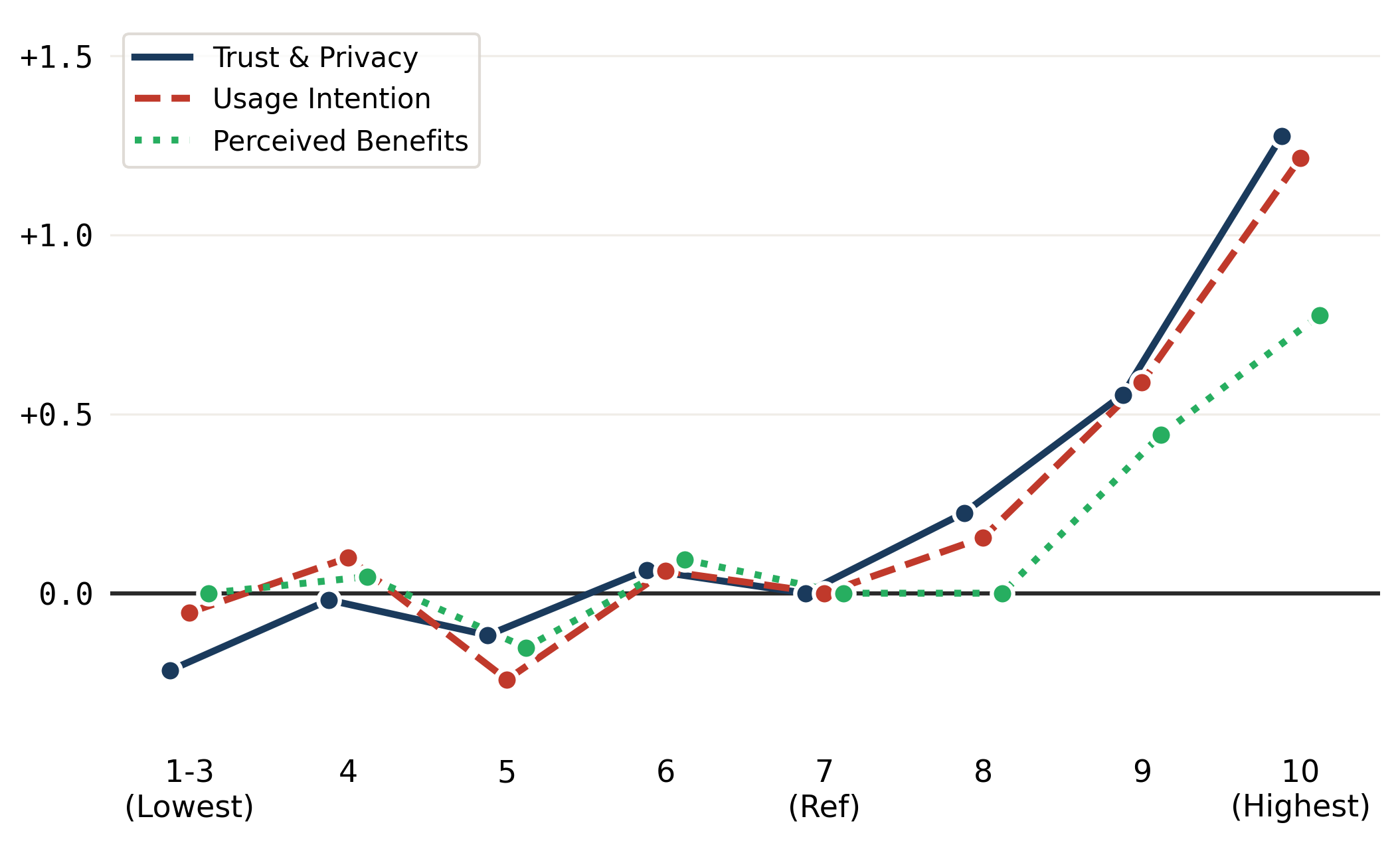}
    \caption{Higher SES drives Trust, Perceived Benefits, and Usage Intention. Fixed effect estimates by Socioeconomic Status (Ref = SES 7).}
    \label{fig:SES_effects}
\end{figure}

We provide the full results table for the CLMMs in this section. Figure~\ref{fig:age_effects} shows the fixed effect estimates by age group, and Figure~\ref{fig:SES_effects} shows the SES estimates across models; the demographics that had the largest negative and positive estimates, respectively. 

The usage intention results (Table~\ref{tab:CLMM_Results_USE}) closely mirror trust and privacy. Age again emerges as a consistent predictor: middle-aged adults (25–44) hold the most positive attitudes, while older adults (55+) show the largest negative effects. Unlike trust and privacy, however, Gender is not significant for usage intention. Among education levels, ``Some college, no degree'' is a significant negative predictor alongside MSc, while High School does not reach significance. Religion and marital status replicate the previous findings: non-religious respondents report lower usage intention, and single individuals score lower than those who were married or partnered, though divorce was not significant. SES remains the most impactful fixed effect, with the highest group (10) again showing the largest positive effect. The random effects tell a similar story: the conditional ICC is 0.050, indicating that only 5.0\% of residual variance is between countries, and the likelihood ratio test again confirms that the country-level effect is significant ($\chi^2(1) = 294.77$, $p < .001$). Country rankings are nearly identical, with Anglophone countries (UK, USA) showing positive intercepts and the Netherlands showing the most negative, reinforcing the cross-construct consistency of cultural differences.

The perceived benefits results (Table~\ref{tab:CLMM_Results_BENF}) largely reinforce the patterns observed for trust and usage. Older adults (55+) again report significantly lower scores. Gender is non-significant, consistent with usage. Religion maintains the same direction across all three constructs: non-religious respondents report lower perceived benefits compared to Christians, while Muslims and other religions do not differ from the reference group. Being single remains a robust negative predictor across all constructs, underscoring the consistent association between marital status and chatbot attitudes. SES remains the most impactful fixed effect: the highest group (10) has the largest positive effect, and groups 9 and above are consistently significant across trust, usage intention, and perceived benefits, confirming that higher socioeconomic status is reliably associated with more favorable chatbot perceptions. The country-level random effects tell a familiar story: the conditional ICC is 0.042, and the likelihood ratio test confirms that between-country variance remains significant ($\chi^2(1) = 330.41$, $p < .001$). The country ranking is virtually identical to trust and usage intention, with the UK and the USA showing positive intercepts and the Netherlands the most negative, further reinforcing the robustness of Anglophone-Continental European attitudinal differences across constructs. Notably, education is the one domain where Perceived Benefits diverged, with no significant effects.

\begin{table}[ht]
    \fontsize{8.5pt}{9pt}\selectfont
    \centering
    \begin{tabular}{@{}lrrrr@{}}
    \toprule
    & Estimate & SE & $z$ & $p$ \\
    \midrule
    \multicolumn{5}{@{}l}{\textit{Fixed Effects}} \\[4pt]
    \multicolumn{5}{@{}l}{\textbf{Gender} (Ref: Male)} \\
    \textbf{\quad Female} & $\mathbf{\text{-}0.117}$ & \textbf{0.034} & $\mathbf{\text{-}3.403}$ & $\mathbf{< .001}$ \\[4pt]
    \multicolumn{5}{@{}l}{\textbf{Age} (Ref: 25-34 years)} \\
    \textbf{\quad 18-24 years} & $\mathbf{\text{-}0.213}$ & \textbf{0.073} & $\mathbf{\text{-}2.925}$ & $\mathbf{.003}$ \\
    \quad 35-44 years & $\text{-}0.051$ & 0.046 & $\text{-}1.101$ & $.271$ \\
    \textbf{\quad 45-54 years} & $\mathbf{\text{-}0.106}$ & \textbf{0.053} & $\mathbf{\text{-}1.994}$ & $\mathbf{.046}$ \\
    \textbf{\quad 55-64 years} & $\mathbf{\text{-}0.253}$ & \textbf{0.066} & $\mathbf{\text{-}3.818}$ & $\mathbf{< .001}$ \\
    \textbf{\quad 65+ years} & $\mathbf{\text{-}0.505}$ & \textbf{0.084} & $\mathbf{\text{-}6.016}$ & $\mathbf{< .001}$ \\[4pt]
    \multicolumn{5}{@{}l}{\textbf{Education} (Ref: BSc)} \\
    \textbf{\quad HS} & $\mathbf{0.144}$ & \textbf{0.055} & $\mathbf{2.616}$ & $\mathbf{.009}$ \\
    \quad AD & $0.056$ & 0.066 & $0.841$ & $.400$ \\
    \quad Some Col. & $\text{-}0.103$ & 0.078 & $\text{-}1.315$ & $.188$ \\
    \quad Prof. Deg. & $\text{-}0.067$ & 0.060 & $\text{-}1.130$ & $.259$ \\
    \textbf{\quad MSc} & $\mathbf{\text{-}0.120}$ & \textbf{0.053} & $\mathbf{\text{-}2.271}$ & $\mathbf{.023}$ \\
    \quad PhD & $0.048$ & 0.083 & $0.578$ & $.563$ \\[4pt]
    \multicolumn{5}{@{}l}{\textbf{Religion} (Ref: Christianity)} \\
    \textbf{\quad None} & $\mathbf{\text{-}0.260}$ & \textbf{0.041} & $\mathbf{\text{-}6.303}$ & $\mathbf{< .001}$ \\
    \quad Islam & $0.135$ & 0.073 & $1.842$ & $.066$ \\
    \quad Other & $0.062$ & 0.090 & $0.686$ & $.493$ \\[4pt]
    \multicolumn{5}{@{}l}{\textbf{Marital status} (Ref: Married/Partnership)} \\
    \textbf{\quad Single} & $\mathbf{\text{-}0.321}$ & \textbf{0.043} & $\mathbf{\text{-}7.509}$ & $\mathbf{< .001}$ \\
    \textbf{\quad Divorced} & $\mathbf{\text{-}0.241}$ & \textbf{0.081} & $\mathbf{\text{-}2.962}$ & $\mathbf{.003}$ \\[4pt]
    \multicolumn{5}{@{}l}{\textbf{SES} (Ref: 7)} \\
    \textbf{\quad 1--3 (Lowest)} & $\mathbf{\text{-}0.215}$ & \textbf{0.088} & $\mathbf{\text{-}2.442}$ & $\mathbf{.015}$ \\
    \quad 4 & $\text{-}0.019$ & 0.070 & $\text{-}0.277$ & $.782$ \\
    \quad 5 & $\text{-}0.117$ & 0.060 & $\text{-}1.949$ & $.051$ \\
    \quad 6 & $0.065$ & 0.054 & $1.198$ & $.231$ \\
    \textbf{\quad 8} & $\mathbf{0.224}$ & \textbf{0.053} & $\mathbf{4.254}$ & $\mathbf{< .001}$ \\
    \textbf{\quad 9} & $\mathbf{0.554}$ & \textbf{0.072} & $\mathbf{7.657}$ & $\mathbf{< .001}$ \\
    \textbf{\quad 10 (Best off)} & $\mathbf{1.275}$ & \textbf{0.084} & $\mathbf{15.239}$ & $\mathbf{< .001}$ \\[6pt]
    \midrule
    \multicolumn{5}{@{}l}{\textit{Random Effects}} \\[4pt]
    \textbf{Country} & \multicolumn{4}{@{}l}{$\sigma^2 = 0.214$, \text{SD} $= 0.463$} \\
    \quad Netherlands & $\text{-}0.542$ & & & \\
    \quad Italy & $\text{-}0.276$ & & & \\
    \quad Spain & $\text{-}0.259$ & & & \\
    \quad Germany & $\text{-}0.208$ & & & \\
    \quad France & $\text{-}0.105$ & & & \\
    \quad United States & $0.581$ & & & \\
    \quad United Kingdom & $0.809$ & & & \\[4pt]
    \midrule
    \multicolumn{5}{@{}p{0.9\columnwidth}}{\tiny HS = High School, AD = Associate Degree, Some Col. = Some college, no degree, Prof. Deg. = Professional Degree, None = No religion.} \\
    \end{tabular}
    \caption{Cumulative Link Mixed Model Results for Trust and Privacy.}
    \label{tab:CLMM_Results}
\end{table}

\begin{table}[htbp]
    \fontsize{8.5pt}{9pt}\selectfont
    \centering
    \begin{tabular}{@{}lrrrr@{}}
    \toprule
    & Estimate & SE & $z$ & $p$ \\
    \midrule
    \multicolumn{5}{@{}l}{\textit{Fixed Effects}} \\[4pt]
    \multicolumn{5}{@{}l}{\textbf{Gender} (Ref: Male)} \\
    \quad Female & $\text{-}0.036$ & 0.041 & $\text{-}0.879$ & $.380$ \\[4pt]
    \multicolumn{5}{@{}l}{\textbf{Age} (Ref: 25-34 years)} \\
    \quad 18-24 years & $\text{-}0.147$ & 0.086 & $\text{-}1.710$ & $.087$ \\
    \quad 35-44 years & $\text{-}0.077$ & 0.054 & $\text{-}1.421$ & $.155$ \\
    \textbf{\quad 45-54 years} & $\mathbf{\text{-}0.147}$ & \textbf{0.063} & $\mathbf{\text{-}2.330}$ & $\mathbf{.020}$ \\
    \textbf{\quad 55-64 years} & $\mathbf{\text{-}0.340}$ & \textbf{0.079} & $\mathbf{\text{-}4.280}$ & $\mathbf{< .001}$ \\
    \textbf{\quad 65+ years} & $\mathbf{\text{-}0.654}$ & \textbf{0.102} & $\mathbf{\text{-}6.394}$ & $\mathbf{< .001}$ \\[4pt]
    \multicolumn{5}{@{}l}{\textbf{Education} (Ref: BSc)} \\
    \quad HS & $0.099$ & 0.066 & $1.498$ & $.134$ \\
    \quad AD & $0.015$ & 0.079 & $0.191$ & $.849$ \\
    \textbf{\quad Some Col.} & $\mathbf{\text{-}0.281}$ & \textbf{0.094} & $\mathbf{\text{-}2.989}$ & $\mathbf{.003}$ \\
    \quad Prof. Deg. & $0.031$ & 0.071 & $0.434$ & $.664$ \\
    \textbf{\quad MSc} & $\mathbf{\text{-}0.194}$ & \textbf{0.062} & $\mathbf{\text{-}3.133}$ & $\mathbf{.002}$ \\
    \quad PhD & $0.181$ & 0.097 & $1.860$ & $.063$ \\[4pt]
    \multicolumn{5}{@{}l}{\textbf{Religion} (Ref: Christianity)} \\
    \textbf{\quad None} & $\mathbf{\text{-}0.282}$ & \textbf{0.049} & $\mathbf{\text{-}5.727}$ & $\mathbf{< .001}$ \\
    \quad Islam & $0.092$ & 0.086 & $1.067$ & $.286$ \\
    \quad Other & $\text{-}0.012$ & 0.106 & $\text{-}0.116$ & $.907$ \\[4pt]
    \multicolumn{5}{@{}l}{\textbf{Marital status} (Ref: Married/Partnership)} \\
    \textbf{\quad Single} & $\mathbf{\text{-}0.262}$ & \textbf{0.051} & $\mathbf{\text{-}5.134}$ & $\mathbf{< .001}$ \\
    \quad Divorced & $\text{-}0.140$ & 0.098 & $\text{-}1.424$ & $.154$ \\[4pt]
    \multicolumn{5}{@{}l}{\textbf{SES} (Ref: 7)} \\
    \quad 1--3 (Lowest) & $\text{-}0.054$ & 0.106 & $\text{-}0.509$ & $.611$ \\
    \quad 4 & $0.099$ & 0.084 & $1.193$ & $.233$ \\
    \textbf{\quad 5} & $\mathbf{\text{-}0.241}$ & \textbf{0.071} & $\mathbf{\text{-}3.411}$ & $\mathbf{< .001}$ \\
    \quad 6 & $0.062$ & 0.064 & $0.976$ & $.329$ \\
    \textbf{\quad 8} & $\mathbf{0.155}$ & \textbf{0.062} & $\mathbf{2.478}$ & $\mathbf{.013}$ \\
    \textbf{\quad 9} & $\mathbf{0.588}$ & \textbf{0.085} & $\mathbf{6.877}$ & $\mathbf{< .001}$ \\
    \textbf{\quad 10 (Best off)} & $\mathbf{1.215}$ & \textbf{0.097} & $\mathbf{12.477}$ & $\mathbf{< .001}$ \\[6pt]
    \midrule
    \multicolumn{5}{@{}l}{\textit{Random Effects}} \\[4pt]
    \textbf{Country} & \multicolumn{4}{@{}l}{$\sigma^2 = 0.173$, \text{SD} $= 0.416$} \\
    \quad Netherlands & $\text{-}0.540$ & & & \\
    \quad Italy & $\text{-}0.346$ & & & \\
    \quad Spain & $\text{-}0.204$ & & & \\
    \quad Germany & $\text{-}0.096$ & & & \\
    \quad France & $\text{-}0.024$ & & & \\
    \quad United Kingdom & $0.585$ & & & \\
    \quad United States & $0.624$ & & & \\[4pt]
    \midrule
    \multicolumn{5}{@{}p{0.9\columnwidth}}{\tiny HS = High School; AD = Associate Degree; Some Col. = Some college, no degree; Prof. Deg. = Professional Degree; None = No religion.} \\
    \end{tabular}
    \caption{Cumulative Link Mixed Model Results for Usage Intention.}
    \label{tab:CLMM_Results_USE}
\end{table}

\begin{table}[t]
    \fontsize{8.5pt}{9pt}\selectfont
    \centering
    \begin{tabular}{@{}lrrrr@{}}
    \toprule
    & Estimate & SE & $z$ & $p$ \\
    \midrule
    \multicolumn{5}{@{}l}{\textit{Fixed Effects}} \\[4pt]
    \multicolumn{5}{@{}l}{\textbf{Gender} (Ref: Male)} \\
    \quad Female & $0.032$ & 0.035 & $0.913$ & $.361$ \\[4pt]
    \multicolumn{5}{@{}l}{\textbf{Age} (Ref: 25-34 years)} \\
    \quad 18-24 years & $0.005$ & 0.073 & $0.075$ & $.941$ \\
    \quad 35-44 years & $\text{-}0.067$ & 0.047 & $\text{-}1.440$ & $.150$ \\
    \quad 45-54 years & $\text{-}0.034$ & 0.054 & $\text{-}0.636$ & $.525$ \\
    \textbf{\quad 55-64 years} & $\mathbf{\text{-}0.255}$ & \textbf{0.067} & $\mathbf{\text{-}3.790}$ & $\mathbf{< .001}$ \\
    \textbf{\quad 65+ years} & $\mathbf{\text{-}0.450}$ & \textbf{0.086} & $\mathbf{\text{-}5.228}$ & $\mathbf{< .001}$ \\[4pt]
    \multicolumn{5}{@{}l}{\textbf{Education} (Ref: BSc)} \\
    \quad HS & $0.017$ & 0.056 & $0.299$ & $.765$ \\
    \quad AD & $\text{-}0.027$ & 0.067 & $\text{-}0.402$ & $.688$ \\
    \quad Some Col. & $\text{-}0.025$ & 0.080 & $\text{-}0.309$ & $.757$ \\
    \quad Prof. Deg. & $\text{-}0.093$ & 0.061 & $\text{-}1.530$ & $.126$ \\
    \quad MSc & $\text{-}0.066$ & 0.053 & $\text{-}1.243$ & $.214$ \\
    \quad PhD & $0.138$ & 0.082 & $1.685$ & $.092$ \\[4pt]
    \multicolumn{5}{@{}l}{\textbf{Religion} (Ref: Christianity)} \\
    \textbf{\quad None} & $\mathbf{\text{-}0.114}$ & \textbf{0.042} & $\mathbf{\text{-}2.717}$ & $\mathbf{.007}$ \\
    \quad Islam & $\text{-}0.037$ & 0.072 & $\text{-}0.514$ & $.607$ \\
    \quad Other & $\text{-}0.030$ & 0.090 & $\text{-}0.338$ & $.735$ \\[4pt]
    \multicolumn{5}{@{}l}{\textbf{Marital status} (Ref: Married/Partnership)} \\
    \textbf{\quad Single} & $\mathbf{\text{-}0.241}$ & \textbf{0.044} & $\mathbf{\text{-}5.542}$ & $\mathbf{< .001}$ \\
    \quad Divorced & $0.009$ & 0.083 & $0.111$ & $.912$ \\[4pt]
    \multicolumn{5}{@{}l}{\textbf{SES} (Ref: 7)} \\
    \quad 1--3 (Lowest) & $0.000$ & 0.090 & $0.002$ & $.998$ \\
    \quad 4 & $0.046$ & 0.071 & $0.649$ & $.516$ \\
    \textbf{\quad 5} & $\mathbf{\text{-}0.152}$ & \textbf{0.061} & $\mathbf{\text{-}2.504}$ & $\mathbf{.012}$ \\
    \quad 6 & $0.094$ & 0.055 & $1.722$ & $.085$ \\
    \quad 8 & $0.000$ & 0.053 & $\text{-}0.005$ & $.996$ \\
    \textbf{\quad 9} & $\mathbf{0.442}$ & \textbf{0.073} & $\mathbf{6.090}$ & $\mathbf{< .001}$ \\
    \textbf{\quad 10 (Best off)} & $\mathbf{0.776}$ & \textbf{0.082} & $\mathbf{9.486}$ & $\mathbf{< .001}$ \\[6pt]
    \midrule
    \multicolumn{5}{@{}l}{\textit{Random Effects}} \\[4pt]
    \textbf{Country} & \multicolumn{4}{@{}l}{$\sigma^2 = 0.143$, \text{SD} $= 0.379$} \\
    \quad Netherlands & $\text{-}0.500$ & & & \\
    \quad Germany & $\text{-}0.243$ & & & \\
    \quad Italy & $\text{-}0.203$ & & & \\
    \quad Spain & $\text{-}0.116$ & & & \\
    \quad France & $\text{-}0.029$ & & & \\
    \quad United States & $0.422$ & & & \\
    \quad United Kingdom & $0.668$ & & & \\[4pt]
    \midrule
    \multicolumn{5}{@{}p{0.9\columnwidth}}{\tiny HS = High School; AD = Associate Degree; Some Col. = Some college, no degree; Prof. Deg. = Professional Degree; None = No religion.} \\
    \end{tabular}
    \caption{Cumulative Link Mixed Model Results for Perceived Benefits.}
    \label{tab:CLMM_Results_BENF}
\end{table}

\clearpage
\section{Prompt Samples and Country Distributions}
\label{PromptSamples}

\begin{figure*}[t]
  \centering
  \includegraphics[width=0.8\textwidth]{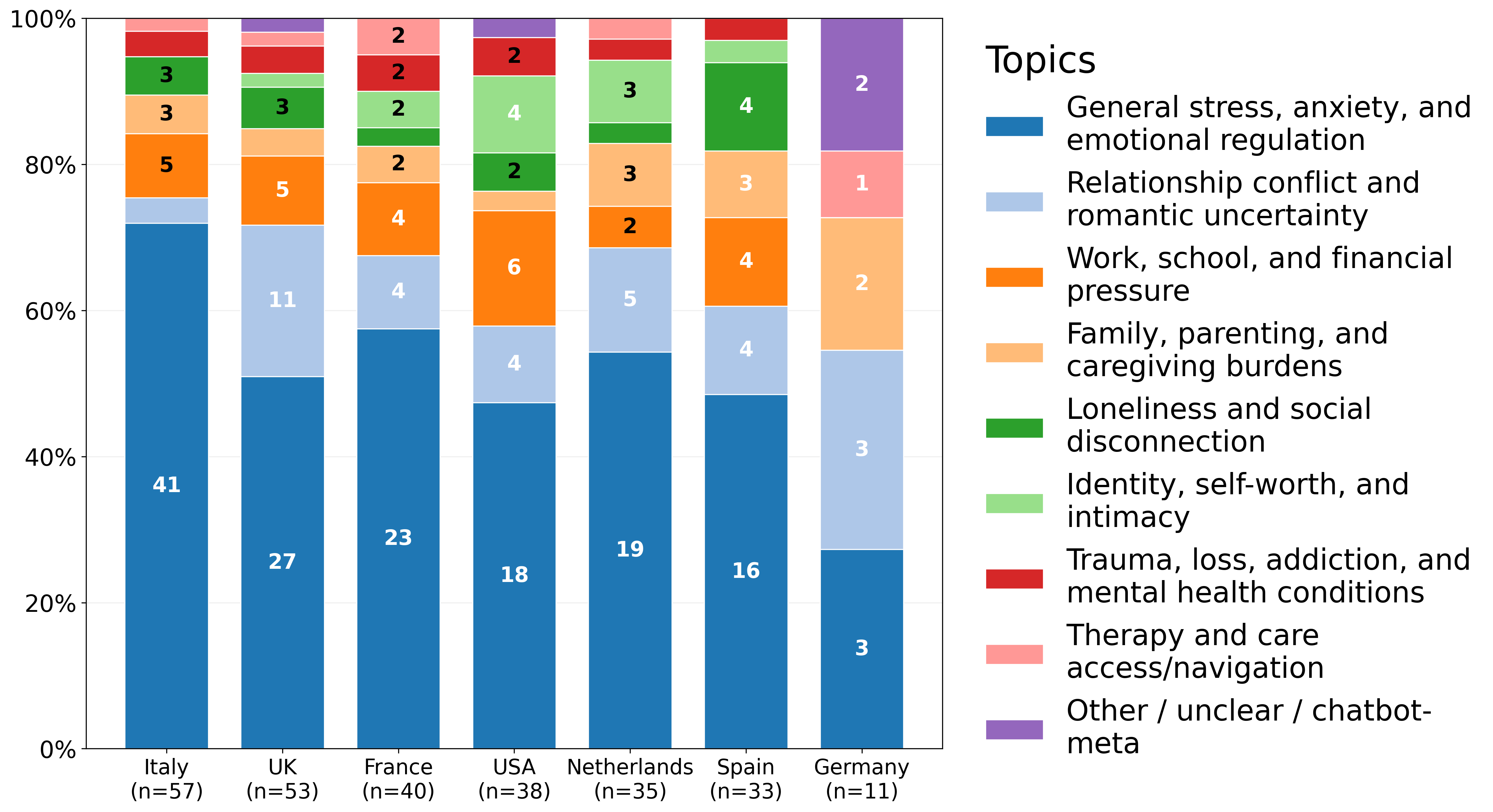}
  \caption{Topic distribution of shared prompts related to mental wellbeing across countries.}
  \label{fig:PromptTopics}
\end{figure*}

Table \ref{tab:prompts_by_country} shows examples of shared prompts representing the following five key topics: general stress and anxiety, relationship conflict, family and caregiving, loneliness, and trauma and mental health.

\begin{table*}[t]
\centering
\small
\renewcommand{\arraystretch}{1.2}
\begin{tabular}{p{3.2cm} p{3.9cm} p{3.9cm} p{3.9cm}}
\toprule
\textbf{Topic} & \textbf{UK} & \textbf{USA} & \textbf{Netherlands} \\
\midrule
General stress \& anxiety
  & ``I'm feeling down today and need some motivational advice or coping strategies to improve my mood.''
  & ``I'm feeling stressed and overwhelmed, can you give me some tips to relax and feel more in control?''
  & ``Welke tips en trucs zijn er om goed om te gaan met stress?'' [What tips are there to deal with stress?] \\
Relationship conflict
  & ``How to handle arguments at home.''
  & ``How to react to the guy I am dating not texting me for a couple of days?''
  & ``Hoe omgaan met wispelturige partner.'' [How to deal with an unpredictable partner.] \\
  Family \& caregiving
  & ``Can you help me with ways I can handle a tiring sibling?''
  & ``What to do when a child upsets me.''
  & ``Hoe moet ik omgaan met het feit dat mijn dochters geen contact willen hebben met mij?'' [How do I cope with my daughters cutting contact with me?] \\
Loneliness
  & ``Most times I feel like I don't belong. Like people always feel off. What could possibly be wrong?''
  & ``How to deal with stress and anxiety alone?''
  & ``Wat te doen tegen eenzaam voelen.'' [What to do about feeling lonely.] \\
Trauma \& mental health
  & ``How to manage stress and depression and overcome an abusive partner.''
  & ``How to deal with a terminal diagnosis.''
  & ``Verwerking rouw dood opa.'' [Processing grief after my grandfather's death.] \\
\bottomrule
\end{tabular}
\caption{Representative shared user prompts across five themes for the UK (country with highest TRU, BENF, and USE construct scores), USA (second highest), and the Netherlands (lowest). Dutch prompts are shown in the original language with translations in brackets. Prompts lightly edited for length and anonymization.}
\label{tab:prompts_by_country}
\end{table*}

\clearpage
\section{Questionnaire}\label{app:quest}

\begin{table*}[tbp]
\centering
\scriptsize
\setlength{\tabcolsep}{5pt}
\renewcommand{\arraystretch}{1.25}
\begin{tabular}{>{\raggedright\arraybackslash}p{2.8cm}
>{\raggedright\arraybackslash}p{8.2cm}
>{\raggedright\arraybackslash}p{3.2cm}}
\toprule
\textbf{Construct} & \textbf{Questionnaire Item} & \textbf{Source} \\
\midrule
\multicolumn{3}{l}{\cellcolor{rowgray}\textbf{Trust}} \\[-2pt]
\rowcolor{rowgray}
& I trust the information or support this chatbot provides.
& \cite{marimon2024} \\
\rowcolor{rowgray}
& The chatbot provides credible information.
& \cite{marimon2024} \\
\rowcolor{rowgray}
& I trust that this chatbot is honest and transparent about what it can and cannot do.
& Authors \\
\rowcolor{rowgray}
& Overall, I trust the chatbot.
& \cite{schmidmaier2024} \\
\midrule
\multicolumn{3}{l}{\textbf{Privacy}} \\[-2pt]
& The chatbot clearly communicates its limitations in providing emotional support.
& Authors \\
& I trust the chatbot to handle sensitive information safely.
& \cite{marimon2024} \\
& I feel confident that my interactions with the chatbot are private.
& \cite{marimon2024} \\
& Overall, I believe this chatbot is safe and respectful of my privacy.
& Authors \\
& I feel safe sharing my emotions with the chatbot.
& \cite{marimon2024} \\
& The chatbot provides advice on when to seek professional assistance or personal support from others.
& Authors \\
\midrule
\multicolumn{3}{l}{\cellcolor{rowgray}\textbf{Perceived Benefits}} \\[-2pt]
\rowcolor{rowgray}
& \textit{Clarity \& Self-Reflection:} The chatbot helps me feel more confident in managing emotional difficulties.
& Authors \\
\rowcolor{rowgray}
& \textit{Challenge \& New Perspectives:} The chatbot challenges my thinking in a constructive way.
The chatbot helps me consider alternative viewpoints on my concerns.
& Authors \\
\rowcolor{rowgray}
& \textit{Recommendations \& Advice:} The chatbot provides recommendations or advice that are useful for my specific situations.
& Authors \\
\rowcolor{rowgray}
& \textit{Actionability:} The chatbot helps me identify actionable steps I can take to improve my mental wellbeing.
& Authors \\
\rowcolor{rowgray}
& \textit{Accessibility:} The chatbot's availability (e.g., 24/7 access) is a significant benefit for my wellbeing support.
& Authors \\
\rowcolor{rowgray}
& \textit{Non-judgmental Exploration:} I feel that I can express my feelings openly without fear of judgment.
& Authors \\
\rowcolor{rowgray}
& \textit{Ability:} Using this chatbot has improved my ability to manage my wellbeing concerns.
& Authors \\
\rowcolor{rowgray}
& \textit{Cost:} The fact that this chatbot is free (or less expensive than traditional therapy) is a major benefit for me.
& Authors \\
\rowcolor{rowgray}
& \textit{General Usefulness:} Overall, I find the chatbot to be a valuable tool for supporting my mental wellbeing.
After using the chatbot, I feel calmer or more relaxed.
& Authors \\
\midrule
\multicolumn{3}{l}{\textbf{Attitude Toward Using}} \\[-2pt]
& Using the chatbot for emotional or mental wellbeing support is a good idea.
& \cite{davis1989} \\
& The chatbot helps me feel emotionally understood.
& \cite{davis1989} \\
& I feel motivated to continue using the chatbot regularly.
& Authors \\
& Overall, my attitude toward using the chatbot is favorable.
& \cite{davis1989} \\
\midrule
\multicolumn{3}{l}{\cellcolor{rowgray}\textbf{Actual System Use}} \\[-2pt]
\rowcolor{rowgray}
& I use the chatbot frequently for wellbeing/emotional support.
& \cite{davis1989} \\
\rowcolor{rowgray}
& My interactions with the chatbot for mental wellbeing or emotional support are a regular part of my routine.
& \cite{davis1989} \\
\rowcolor{rowgray}
& When I need mental wellbeing or emotional support, the chatbot is usually one of the first resources I turn to.
& \cite{davis1989} \\
\bottomrule
\end{tabular}
\caption{Survey Constructs, Items, and Sources. Authors: Items proposed by authors.}
\label{tab:survey_items}
\renewcommand{\arraystretch}{1}
\end{table*}

\end{document}